\documentclass[11pt]{article}

\usepackage[utf8]{inputenc}
\usepackage[T1]{fontenc}
\usepackage{times}

\usepackage{amssymb, amsmath}
\usepackage{hyperref}
\usepackage{url}
\usepackage{tabularx}
\usepackage{adjustbox}
\usepackage[ruled,vlined,algo2e]{algorithm2e}
\usepackage[capitalise]{cleveref}
\usepackage{float}
\usepackage[numbers,sort&compress]{natbib}

\usepackage{xcolor}

\title{\texttt{Pruning AMR}: Efficient Visualization of Implicit Neural Representations via Weight Matrix Analysis}

\author{
  Jennifer Zvonek\thanks{Center for Applied Mathematics, Cornell University, Ithaca, NY 14853, USA. Corresponding author: \texttt{jez34@cornell.edu}.}
  \and
  Andrew Gillette\thanks{Center for Applied Scientific Computing, Lawrence Livermore National Laboratory, Livermore, CA 94550, USA. \texttt{gillette7@llnl.gov}.}
}

\date{}  

\begin{document}
\maketitle

\begin{abstract}
An implicit neural representation (INR) is a neural network that approximates a spatiotemporal function.  
Many memory-intensive visualization tasks, including modern 4D CT scanning methods, represent data natively as INRs.  
While INRs are prized for being more memory-efficient than traditional data stored on a lattice, many visualization tasks still require discretization to a regular grid.
We present \texttt{PruningAMR}, an algorithm that builds a mesh with resolution adapted to geometric features encoded by the INR.  
To identify these geometric features, we use an interpolative decomposition pruning method on the weight matrices of the INR.
The resulting pruned network is used to guide adaptive mesh refinement, enabling automatic mesh generation tailored to the underlying resolution of the function.  
Starting from a pre-trained INR---without access to its training data---we produce a variable resolution visualization with substantial memory savings.
\end{abstract}


\section{Introduction and Motivation}
\label{sec:intro}
Implicit neural representations (INRs) have gained attention in recent years for their ability to represent spatial and time-varying spatial data efficiently. 
While INRs are best known for their fast and accurate visualization applications, these methods only apply to specific neural graphics primitives---such as signed distance functions---and require training routines and data structures---such as hashing techniques---to realize interactive visualization.
For INRs encoding data not derived from graphics primitives, the recourse for visual analysis is to discretize the INR to a uniform grid, thereby enabling traditional techniques, but eliminating any computational savings afforded by the INR encoding. 
This presents an open challenge to communities using INRs in new contexts: given a pre-trained INR, how can we visuallize the encoded information efficiently?
  
The need for efficient visualization of INR data is evidenced by emergent ``dynamic micro-CT'' technology for additive manufacturing.
Recently developed methodologies are capable of storing time-varying volumetric data of materials undergoing physical changes as an INR with $(x,y,z,t)$ inputs. 
In one example, the size of an INR checkpoint file is on the order a few megabytes, but the potential resolution of the time-varying volume is $1024\times1024\times1024\times700$, roughly 3 \textit{terabytes} worth of data in a uniform discretization, well beyond the capabilities of common visualization software.  
Visual inspection of time slices shows that many regions of the domain vary little, whereas some require full resolution for evaluation by subject-matter experts.
Hence, an approach to adaptively sample the INR in a way that preserves fine-grained geometric features of the function is of practical interest, with immediate benefits to dynamic micro-CT technology.

We present \texttt{PruningAMR}, an algorithm that visualizes a pre-trained INR on an adaptive mesh, achieving accuracy comparable to a uniform mesh but with a reduced memory footprint.
\texttt{PruningAMR} begins with a coarse uniform mesh of the domain and iteratively refines elements in which the INR encodes finer-scale features.
We assume knowledge of the INR architecture, as would be included in a standard checkpoint file, but do not assume access to any training data; refinement deciscions are based solely on the weight matrices of the INR.

The refinement decision for a given element is based on the outcome of a ``pruning'' method applied to the weight matrices of the INR, restricted to the element's domain. 
Elements that admit substantial pruning with only a small loss of accuracy are presumed to exhibit low-rank structure and are deemed sufficiently resolved.
Conversely, elements for which significant pruning is not possible, or for which pruning causes significant information loss, are flagged for refinement.

In this paper, we combine traditional adaptive mesh refinement methods with modern neural network compression techniques to visualize data encoded by INRs. The resulting contributions from our work include:
\begin{enumerate}
  \item A general purpose measure of the complexity of geometric features encoded by an INR.
  \item An algorithm to drive adaptive mesh refinement based on assessed complexity of geometric features, as determined by analysis of weight matrices, \emph{without access to the INR’s training data}.
  \item Proof-of-concept demonstrations with quantified memory savings on INRs in 2, 3, and 4 dimensions.
  \item Demonstration of the algorithm on a physics-informed neural network (PINN), a type of INR of significant interest to modeling and simulation communities.
  \item Application of the algorithm to an INR trained on 4D (space + time) data from a state-of-the-art CT scanner.
\end{enumerate}

\section{Background and Literature Comparison}
\subsection{Implicit Neural Representations}
An implicit neural representation (INR) is a type of neural network that approximates a scalar- or vector-valued field with inputs representing physical space or spacetime coordinates. 
The original use of INRs in the context of visualization was to efficiently store an implicit representation of an image~\citep{sitzmann2020implicit}, but interest in the technique quickly grew to include volumetric visualizations as well~\citep{mildenhall2021nerf}.
The output of the popular physics-informed neural network (PINN) for approximating solutions to partial differential equations is a coordinate-valued, multi-layer perceptron (typically), and hence also qualifies as an INR~\citep{karniadakis2021physics}.

The appeal of INRs over traditional discretization is the network's ``ability to model fine detail that is not limited by the grid resolution but by the capacity of the underlying network architecture'' ~\citep{sitzmann2020implicit}.
Only the weights and biases of the INR need to be stored in order to recover the value of the field at the highest level of detail anywhere in the represented domain. 
Accordingly, the INR data structure takes up orders of magnitude less storage than an equivalent standard representation. 
Still, the savings in data storage come with a tradeoff: evaluating the INR can only be done ``pointwise'', meaning discretization and interpolation over a fixed grid of some type is required to employ standard visualization software for all but very specific types of INR data.

\subsection{Visualization and discretization}

While our work is related to both visualization using INRs and traditional data discretization methods, neither of the associated research communities offers a solution to the problem addressed here. 
Much of the visualization work on INRs focuses on methods to train INRs more efficiently, such as ACORN~\citep{martel2021acorn}, scene representation networks~\citep{wurster2023adaptively}, and Instant-NGP~\citep{wurster2023adaptively}.
None of these works, however, addresses the question of how to process, analyze, or efficiently visualize a pre-trained INR without resorting to storing a dense, uniform grid of data.
A separate body of work looks at efficient management and visualization of data stored on adaptive meshes, such as multi-functional approximation~\citep{peterka2023towards}, CPU ray tracing~\citep{wang2020cpu}, and p4est~\citep{burstedde2011p4est}.
These works presume data is provided on an adaptive mesh as input to their use cases, rather than as a pre-trained INR.

We treat INRs as a native data format, akin to a compressed version of a much larger dataset.
The input to our method is a user-provided INR, with no access to the training data.
As output, we produce an adaptive mesh on which the INR has been sampled at vertices, allowing subsequent visualization and analysis via established techniques for meshes.

\subsection{Adaptive Mesh Refinement (AMR)}
Adaptive mesh refinement (AMR) is often thought of in the context of finite element methods (FEM).  
There is a rich field of AMR techniques for FEM incorporating neural networks, from very early ideas based on estimation of a ``mesh density function''~\cite{chedid1996automatic,dyck1992determining} to the deployment of modern reinforcement learning packages for guiding refinement decisions~\cite{dzanic2024dynamo,gillette2024learning}.

While the algorithm and results presented in this work may appear at first to fall into the realm of AMR for FEM, the problem domain here is fundamentally different.
Our algorithm visualizes an INR and does not assume there is any underlying differential equation (although it is possible that there exists one associated with the INR). 
In the CT INR examples studied here, for instance, the function encoded by the INR is not meant to be the approximate solution to a partial differential equation (PDE).  
Since AMR for FEM techniques employ some type of PDE-based error estimator, they cannot be viewed as alternatives to the method we provide here.

\subsection{Pruning via interpolative decomposition of weight matrices}
\label{subsec:pruningID}
The term \emph{pruning} refers to the process of selectively removing weights and biases from a neural network in a way that preserves its mapping from inputs to outputs; see, e.g.~\citep{lee2018snip,li2016pruning,liebenwein2019provable,liu2018rethinking,mussay2019data}.
We use the pruning method proposed by~\citet{IDPruning}, which identifies and merges neurons whose activations can be approximated by a linear combination of the activations of other neurons in the same layer.
The method for detection of such neurons employs a structured low-rank approximation called an \emph{interpolative decomposition (ID)}.
We selected this pruning method due to its theoretical guarantees, ease of implementation, and few number of hyperparameters.

We define notation before describing the ID pruning method.
In this work, we only consider INRs that consist of fully-connected linear layers.~\footnote{Similar pruning methods exist for other layer types and could be substituted for the pruning portion of our algorithm.}
Each layer takes a batch of $\ell$ inputs $x \in \mathbb{R}^{\ell \times n}$, produces an output $y \in \mathbb{R}^{\ell \times m}$, and has a corresponding weight matrix $W \in \mathbb{R}^{m \times n}$ and bias vector $b \in \mathbb{R}^{\ell \times m}$ (where each column of $b$ is identical). 
We assume that the output of the layer is computed as 
\begin{equation*}
  y = g( x W^T + b),
\end{equation*}
where $g$ is the activation function used for the layer. 

We explain the method of ID pruning using a single layer neural network; the extension to multilayer networks is straightforward.
An ID of $W$ is a decomposition of the form $W \approx W_{:, \mathcal{I}} D$, where $\mathcal{I} \subseteq \{1, 2, ..., m\}$, $|\mathcal{I}| = k$, and $D \in \mathbb{R}^{k \times n}$ is called the \emph{interpolation matrix}.
Suppose the neural network has a single hidden layer with output layer weight matrix $U$ and output layer bias vector $c$. 
Then the output of the network $N$ with input $x$ is 
\begin{equation*}
  N(x) := Z(x) U^T + c, 
\end{equation*}
where 
\begin{equation*}
  Z(x) = g(xW^T + b)
\end{equation*}
is the output of the hidden layer.

Let $Z(X) \approx Z_{:, \mathcal{I}} D$ be an ID of $Z(X)$. 
We then have
\begin{align*}
  Z(X) &\approx Z_{:, \mathcal{I}} D \\
  &= g(X W^T + B)_{:, \mathcal{I}} D \\
  &= g(X (W_{\mathcal{I}, :})^T + B_{\mathcal{I}}) D.
\end{align*}
Thus, the output of the full network with a pruned hidden layer is 
\begin{align*}
  N(X) &= g(X (W_{\mathcal{I}, :})^T + B_{:, \mathcal{I}}) D U^T + C \\
  &= g(X \bar{W}^T + \bar{B}) \bar{U}^T + C,
\end{align*}
where we define $\bar{W} := W_{\mathcal{I}, :}$, $\bar{B} := B_{:,  \mathcal{I}}$, and $\bar{U} := U D^T$ to be the new weights and biases of the pruned network. 
Thus, pruning a layer not only affects the weights and bias of that layer, but also the weights of the following layer. 
Note that pruning a layer to rank $k$ results in a layer with $k$ neurons.
Accordingly, the next layer's weights must be updated to accept the new, smaller number of inputs coming from the current layer. 

Given $\varepsilon>0$, the goal of ID pruning is to find $\mathcal{I}$ and $D$ such that $\|W - W_{:, \mathcal{I}} D \|_2 \leq \varepsilon \|W\|_2$, with $|\mathcal{I}|$ as small as possible. 
We use the rank-revealing QR factorization approach from~\citet{IDPruning} to find this decomposition.
For neural networks with more than one hidden layer, IDs for each layer's weight matrix can be computed in parallel. 
However, because the pruned weights of a layer depend on the ID of the previous layer, the final weights of the pruned network must be determined sequentially starting from the ID of the first layer, then the ID of the second layer, and so forth.

\section{The~~\texttt{PruningAMR}~~Algorithm} \label{sec:algorithm}

At a high level, the \texttt{PruningAMR} algorithm is an adaptive refinement method. 
As inputs, it requires the weights and activation functions of an INR defined over a box in $\mathbb{R}^n$ and a handful of hyperparameter choices.  
Starting from a coarse mesh of the box domain, the algorithm passes over each element in the mesh and marks whether or not to refine the element further.  
When all elements are marked, the geometric refinement (subdivision) is carried out and the analysis repeats for all the newly created, smaller mesh elements. 
This continues until a stopping criteria is met, usually achieved by reaching a user-specified maximum number of refinement levels.
The final output is the adaptive rectilinear mesh when the stopping criteria is attained, along with the values of the INR at the vertices of the final mesh.

\begin{figure}
\[
\includegraphics[width=\linewidth]{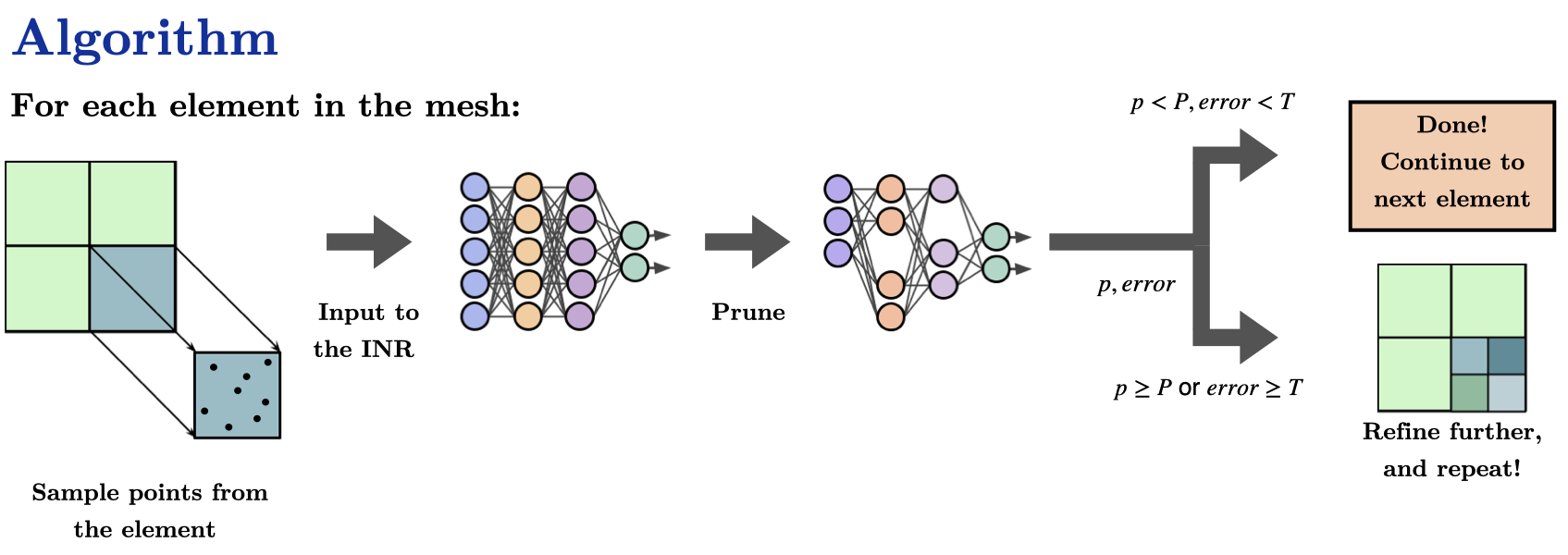} 
\]
\caption{Graphical depiction of an iteration of the \texttt{PruningAMR} algorithm. The algorithm takes as input an initial mesh, a pre-trained INR, an error tolerance $T$, and a proportion $P \in (0,1)$ of the maximum number of neurons allowed after pruning to end refinement for an element. For each element in the mesh, we sample a random set of points from that element. We then pass these points into the INR and use the output to decide how to prune the INR. After pruning, we record the proportion $p$ of remaining neurons to the original number of neurons and the relative error ($error$) of the pruned INR restricted to that element. We compare these values to $P$ and $T$ to decide whether or not to refine the element. We repeat this procedure until all elements satisfy $p<P$ and $error < T$ or the maximum number of iterations is reached.}
\label{fig:algorithm}
\end{figure}

Figure~\ref{fig:algorithm} illustrates the refinement decision step of the algorithm, highlighting the INR-specific analysis that guides where to refine.
Given a mesh element $E$, the original INR is pruned using a set of points sampled uniformly from $E$, following the interpolative decomposition pruning method described in Section~\ref{subsec:pruningID}.
The result of this pruning is two quantities: $p$, the proportion of neurons in the pruned INR to the neurons in the original INR; and $error$, the relative error of the pruned INR over $E$.
The value of $error$ is computed by sampling an additional set of points over $E$ and computing the mean relative error between the pruned INR and original INR.
If $p$ and $error$ are both smaller than user-specified thresholds, the element is not marked for further refinement. 
If either threshold is exceeded, the element is marked.

A few remarks are worth noting to avoid common points of confusion.
At any stage of the algorithm, the pruning is carried out only to determine whether to refine; the pruned networks are not saved or used again.  
Pruning always begins from the original INR, but is performed on the INR restricted to the geometric subdomain defined by the element in consideration.
More precisely, a batch of points are sampled uniformly from the element's domain and then passed to the original INR to inform pruning decisions. 
This allows the INR to be pruned to contain only the neurons necessary to represent the underlying function on the element's domain.

At the conclusion of the algorithm, we compute and store the value of the INR at each vertex of the adaptive mesh to pass to visualization software.
Tools such as Paraview~\citep{Paraview} and~\citet{glvis-web} can robustly interpolate and visualize functions defined only at the vertices of adaptive rectilinear grids. 
We use both of these tools for the visualizations in this work.

A formal description of the algorithm is provided in Algorithm~\ref{alg:amr4inr}. 
We use the notation \texttt{prune($f_\theta$,$E$,$\varepsilon$,$n_{ID}$)} to denote a call to an interpolative decomposition pruning method as described in Section~\ref{subsec:pruningID}.
The inputs to \texttt{prune} are a pre-trained INR $f_\theta$, a mesh element $E$, an error limit for the interpolative decomposition $\varepsilon$, and the number of samples $n_{ID}$ to use when computing an interpolative decomposition.
We use $\texttt{random}(n_{err}, \texttt{domain}(E))$ to denote the generation of $n_{err}$ points drawn uniformly from the geometric domain defined by $E$.

% algorithm here
\begin{algorithm2e}
  \SetKwInOut{Input}{input}
  \SetKwInOut{Output}{output}

  \caption{\texttt{PruningAMR} Algorithm: using adaptive mesh refinement to find a memory-efficient visualization of an INR.}

  \Input{INR $f_\theta$ with set of neurons $\theta$,\\
         inital mesh $M$, \\
         error threshold $T$, \\
         proportion threshold $P$, \\
         interpolative decomposition error limit $\varepsilon$, \\
         maximum number of iterations $K_{max}$, \\
         number of samples for error check $n_{err}$, \\
         number of samples to use for ID $n_{ID}$.}

  \For{$i = 1$ \KwTo $K_{max}$}{
    \For{each element $E$ in $M$ with $\texttt{done\_refining}(E)$ == \texttt{False}}{
      // prune INR on element E\\
      $f^{pruned}_{\tilde{\theta}} \leftarrow \texttt{prune} (f_\theta, E, \varepsilon, n_{ID})$\;
      ~\\
      // compute proportion of neurons remaining after pruning\\
      $p \leftarrow |\tilde{\theta}| / |\theta|$ \;
      ~\\
      // sample random points in domain of E\\
      $X    \leftarrow \texttt{random}(n_{err}, \texttt{domain}(E))$\;
      ~\\
      // compute mean relative error\\
      $error \leftarrow \texttt{mean}\left(|f_\theta(X)- f^{pruned}_{\tilde{\theta}}(X)| / |f_\theta(X)| \right)$ \;
      ~\\
      // Refine all elements that exceed the error or proportion threshold\\
      \If{$error > T$ \texttt{or} $p > P$}{
        $\texttt{refine}(E)$\;
      }
      \Else{
        $\texttt{done\_refining}(E) \leftarrow \texttt{True}$ \;
      }
    }
  }
  \Output{Refined mesh $M$ and values of $f_\theta$ at vertices of $M$}
  \label{alg:amr4inr}
\end{algorithm2e}

The main inputs of Algorithm~\ref{alg:amr4inr} are the INR $f_\theta$, the initial mesh $M$, and the stopping criteria thresholds $T$ and $P$.
The remaining inputs are effectively hyperparameters that can be set heuristically as follows.
The interpolative decomposition error limit $\varepsilon$ is determined by the resolution of the problem; we have found it only needs to be correct up to an order of magnitude. We found $\varepsilon \in \{10^{-2}, 10^{-3}\}$ worked well in our testing, with the choice dependent on the problem dimension and desired accuracy. We recommend starting with $\varepsilon = 10^{-3}$ and adjusting.
The maximum number of iterations $K_{max}$ is determined by the memory and/or time limits of the compute resource; we used at most $K_{max}=8$. 
The number of samples for error computation $n_{err}$ can be $O(100)$ for problems in dimensions 2--4.
We found that past some point, increasing $n_{err}$ does not improve the quality of the result since the error estimate is eventually accurate enough for the purposes of the algorithm.
The number of samples $n_{ID}$ can be set as the width of the layers of the INR since this is sufficient to carry out the rank-based analysis of the interpolative decomposition.

% cost of ID pruning on an element
The computational complexity of Algorithm~\ref{alg:amr4inr} is primarily driven by the complexity of the ID pruning routine.
Recall that the ID pruning method computes QR decompositions of the output matrices from each layer, each of which is $w \times n_{ID}$, where $w$ is the width of the layer. 
Given an INR with $h$ hidden layers of width $w$, we may assume $n_{ID} \propto w$, per our empirically tested heuristic of choosing $n_{ID}$ equal to the width of a layer.
We use LAPACK to compute a rank-revealing QR decomposition of the output matrices; the complexity of this routine for a matrix $A \in R^{m\times n}$ is $\mathcal{O}(mnk)$, where $k = \min(n,m)$~\citep{LAPACK}.
Hence, the complexity of pruning the INR once is $\mathcal{O}(hw^3)$.
Now, for each iteration of Algorithm~\ref{alg:amr4inr}, we call the ID pruning routine once per element not marked \texttt{done\_refining}.
It is not possible \textit{a priori} to estimate how many elements will be marked in a given iteration, limiting a general estimate of algorithmic complexity.
However, calling the ID pruning routine on all elements in a mesh at a fixed iteration can be done in an
embarrassingly parallel fashion, meaning the total compute time could be driven down to $\mathcal{O}(hw^3)$ given sufficient compute resources.

\begin{figure}[h!]
\centering
\begin{tabular}{ccl}
\includegraphics[width=0.37\linewidth]{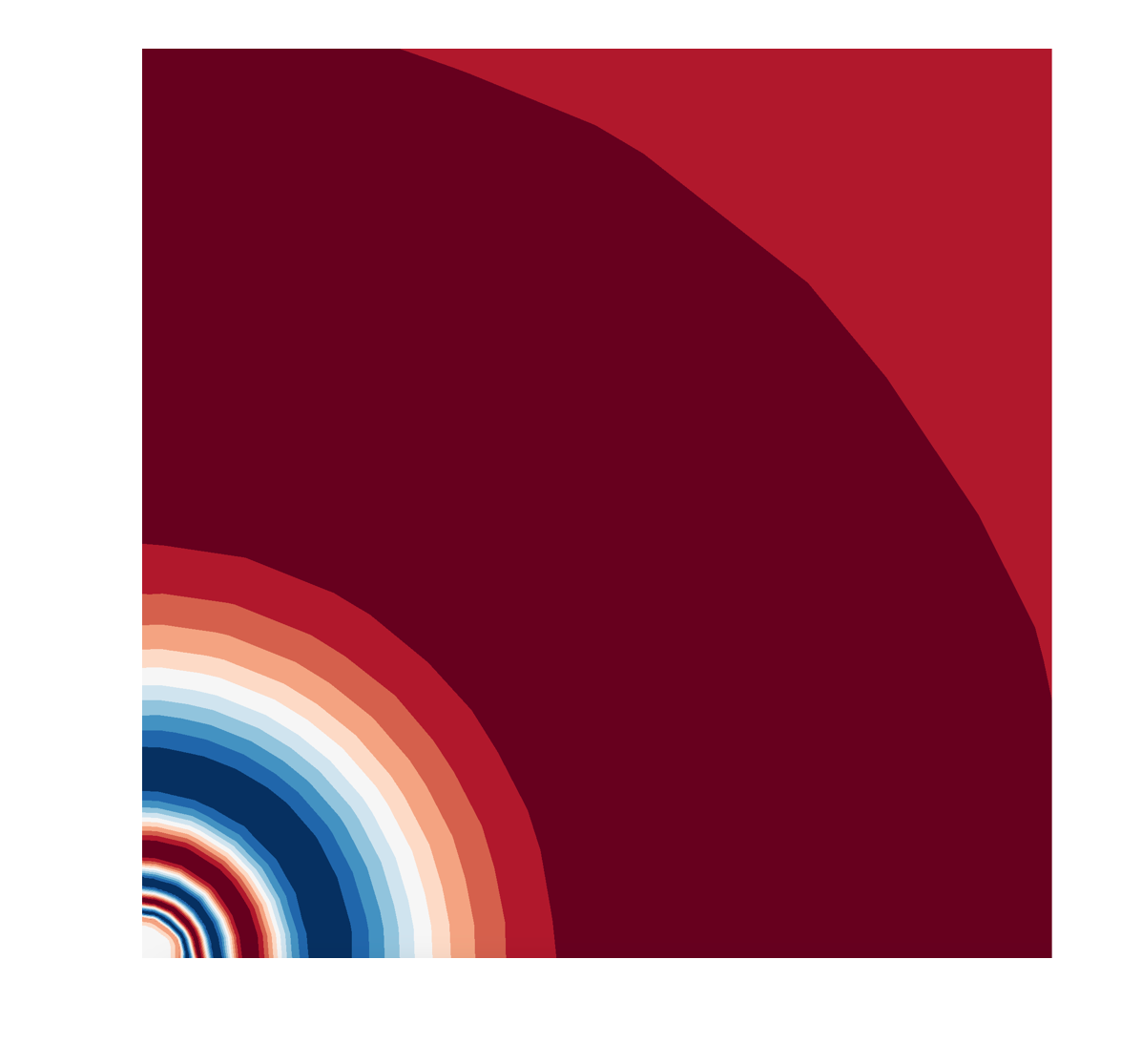}
& \includegraphics[width=0.37\linewidth]{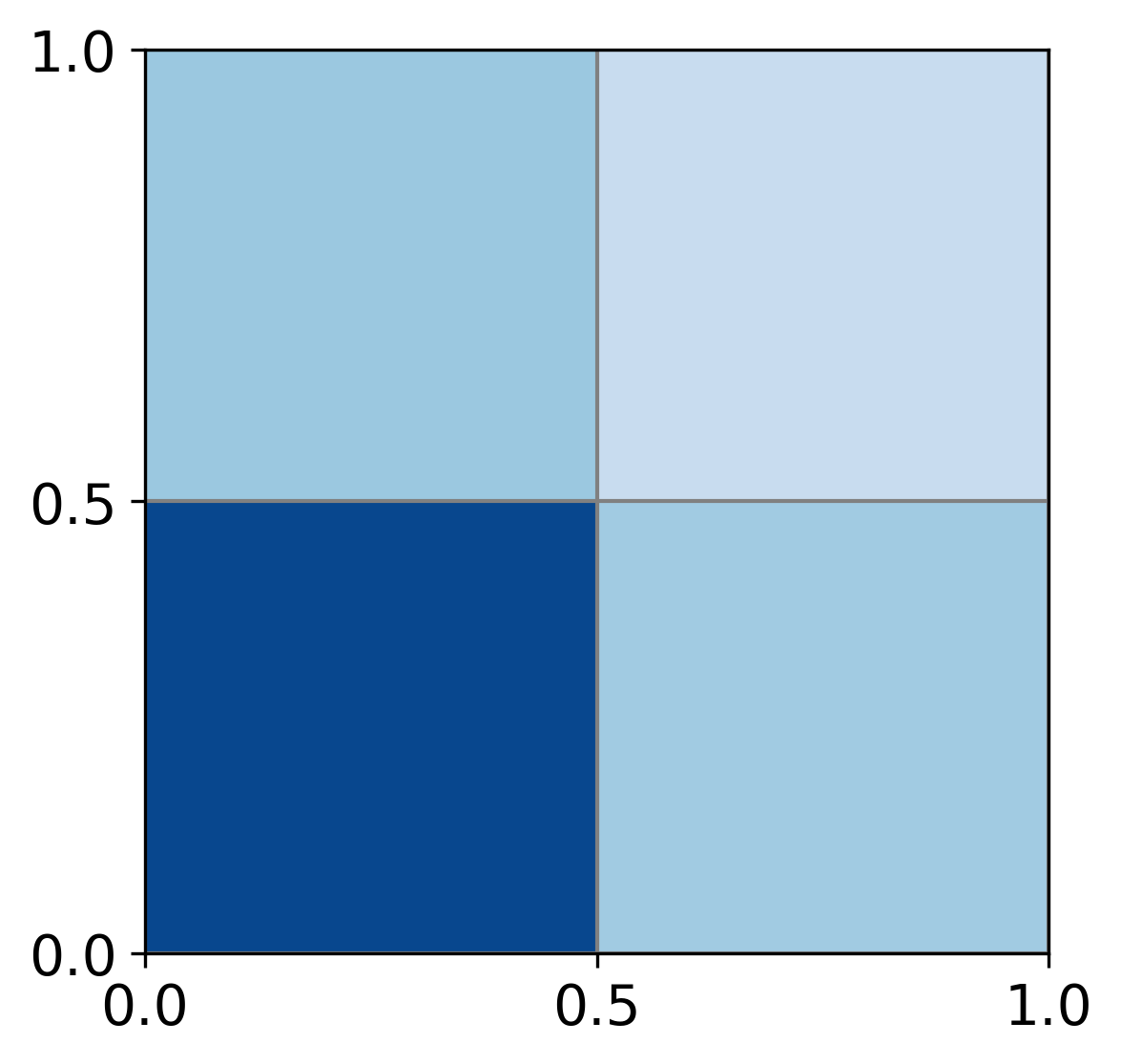} 
& \includegraphics[width=0.1\linewidth]{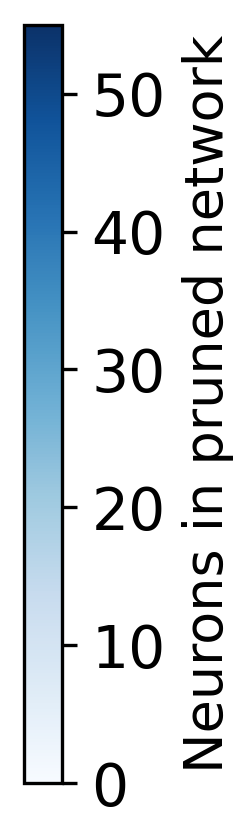} \\
\includegraphics[width=0.37\linewidth]{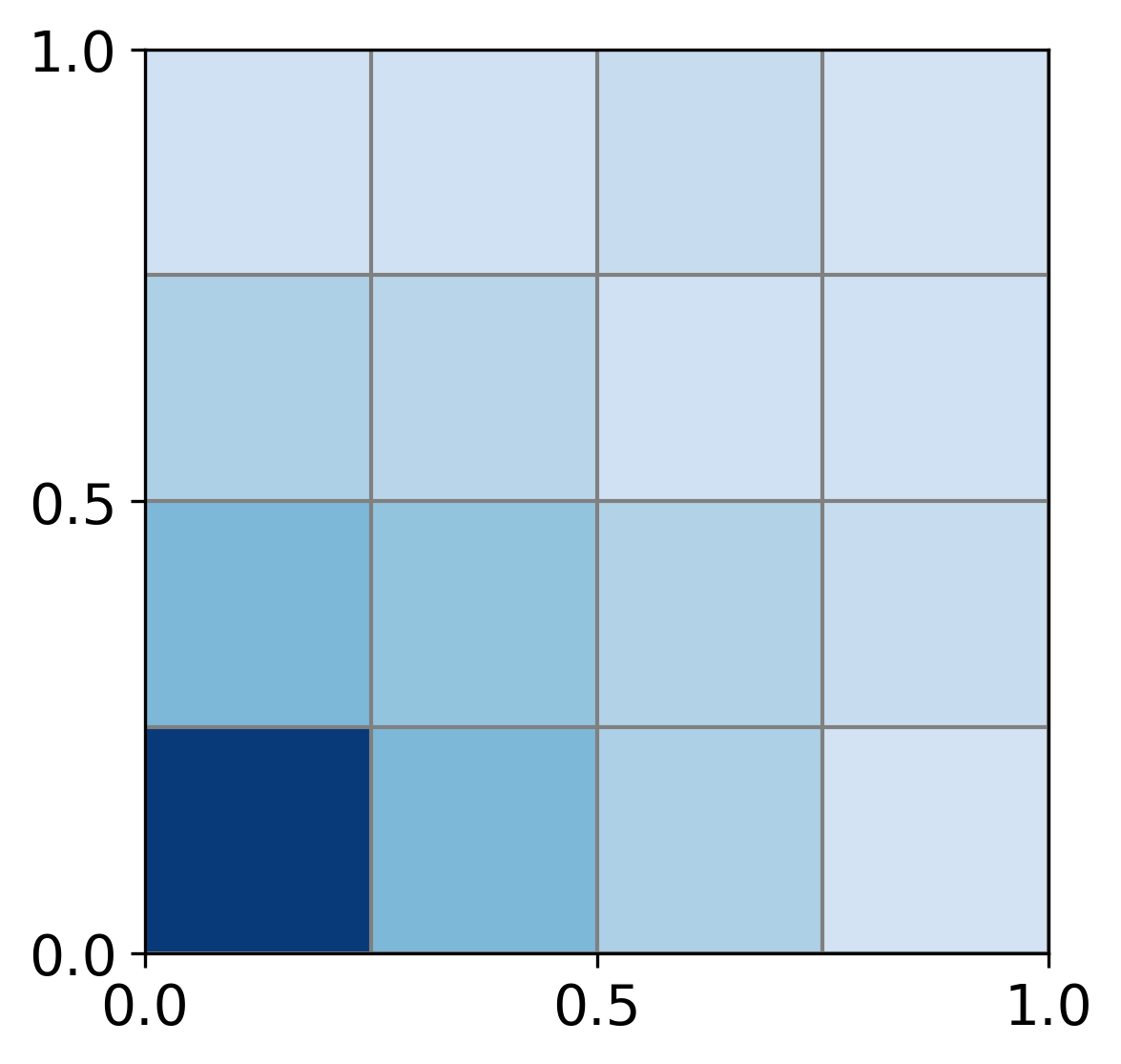}
&  \includegraphics[width=0.37\linewidth]{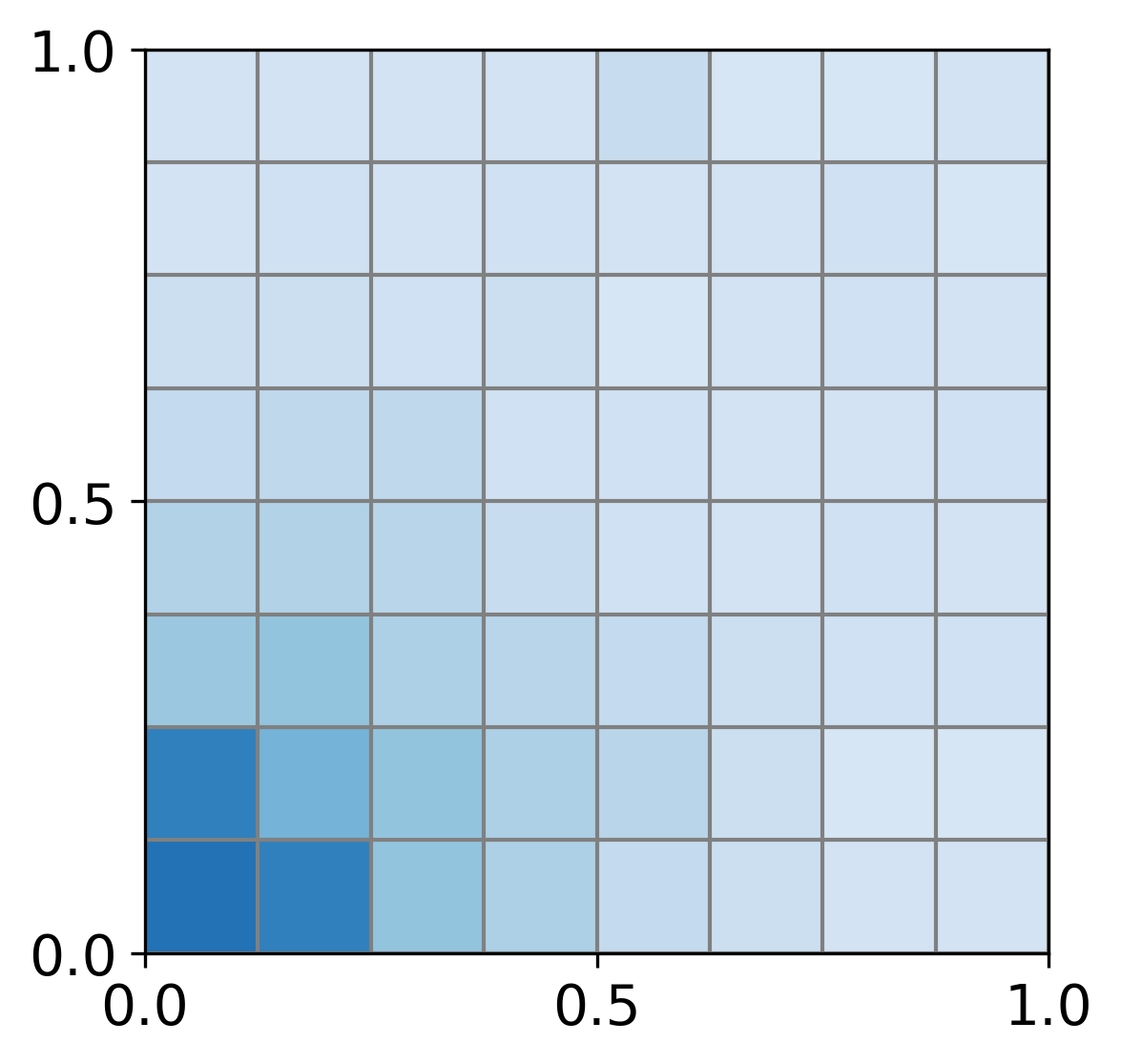} 
& \includegraphics[width=0.1\linewidth]{neuron_count_colorbar.png} \\
\end{tabular}
\caption{As validation for the approach of Algorithm~\ref{alg:amr4inr}, we use an INR trained to fit the 2D oscillatory function shown in the upper left.
On each element in each of the three meshes shown, we call the \texttt{prune} routine, keeping $\varepsilon$ and $n_{ID}$ fixed, and report the number of neurons at the conclusion of pruning.
Notice that more neurons are needed near the origin, where the function has higher oscillation. Also notice that as the elements get smaller with each respective mesh, fewer neurons may be needed per element to describe a given region of the domain.
These findings are consistent with the notion that pruning to a smaller neuron count correlates with simpler geometric features present in that region.}
\label{fig:neuron_counts}
\end{figure}

In Figure~\ref{fig:neuron_counts}, we provide a basic validation check of the code and its underlying motivation as illustrated by a simple example.
We trained an INR to encode a 2D function that is highly oscillatory near the origin (see the top left plot in Figure~\ref{fig:neuron_counts}).
We then fixed values for $\varepsilon$ and $n_{ID}$ and called the \texttt{prune} routine on each element in the mesh. 
For each element, we recorded the number of neurons at the termination of pruning. 
We show the results on three uniformly refined meshes of increasing resolution.

In Figure~\ref{fig:neuron_counts}, darker colors indicate larger neuron counts after pruning.
In all three meshes, the elements closest to the origin are the least pruned, reflecting the higher oscillation of the function in that region of the domain.
Meanwhile, the portion of the function near the top right corner of the domain is much smoother and can be represented by a neural network with fewer neurons. 
We use this observation to motivate our choice of pruning for identifying which regions of the INR's domain have complex features that need higher resolution to depict.

Furthermore, notice that as the mesh is refined further, we often need fewer neurons per element than in the coarser mesh, even in the highly oscillatory region. In particular, the element with a vertex at the origin needs fewer neurons in the third mesh than in the first one. 
This observation tells us that if we keep refining a region, we expect to reach a resolution at which a small neural network can sufficiently describe the function restricted to an element's domain. 
Thus, it is reasonable to refine until the neuron proportion threshold and error threshold are met; these criteria correspond to accurately representing the corresponding feature.
For a further exploration of this 2D example, see Section~\ref{sec:2D_example}.

We implemented Algorithm~\ref{alg:amr4inr} and released it publicly at \url{https://github.com/LLNL/PruningAMR}~\citep{PruningAMR_code}.

\section{Results} \label{sec:Results}
In this section we demonstrate the effectiveness of \texttt{PruningAMR} on a few examples in different dimensions. 
We consider two alternatives to \texttt{PruningAMR} for comparison: \texttt{Uniform} refinement and \texttt{BasicAMR}. 
The \texttt{Uniform} method carries out refinement on every element until a maximum number of iterations is reached. 
The \texttt{BasicAMR} method is a simplified form of \texttt{PruningAMR} that uses no pruning. 
\texttt{BasicAMR} also performs AMR. However, instead of using pruning as a heuristic for where to refine, \texttt{BasicAMR} uses an estimate of the error in each mesh element to decide whether to refine. 
The error estimate for an element is given by taking $n_{err}$ randomly chosen points in the element and computing the mean relative error between $f_\theta$ and the multilinear interpolation of the vertex data.
If the error estimate is larger than a threshold $\tau$, \texttt{BasicAMR} marks the element for refinement.
Note that the relative error computed in \texttt{BasicAMR} is distinct from the relative error of the pruned INR computed in \texttt{PruningAMR}~\ref{alg:amr4inr}; hence, $\tau$ is a distinct hyperparameter from $T$.
We compare with \texttt{BasicAMR} to demonstrate the utility of using pruning for deciding where to refine.

To assess the effectiveness of a refinement method quantitatively, we record the number of degrees of freedom (DOFs)---meaning the number of vertices in the mesh---and an approximation of the total error at each iteration.
The total error at a given iteration is approximated as follows:  
First we sample a large number of points randomly (with uniform distribution) across the entire INR domain.
At each point, we compute the value of the true INR and the bilinear interpolant of the mesh element containing that point, using the true INR values at the element vertices.
The root mean squared error across all sample points is then recorded as the total error.
Plotting error versus DOFs is standard practice in analysis of adaptive mesh refinement schemes. Thus, we include error verus DOFs plots for each of the following examples.

\subsection{2D Validation Example: Analytical oscillation at a corner} \label{sec:2D_example}

\begin{figure}[h]
    \centering
    % Left: Image
    \begin{minipage}[c]{0.45\textwidth}
        \centering
        \includegraphics[width=\linewidth]{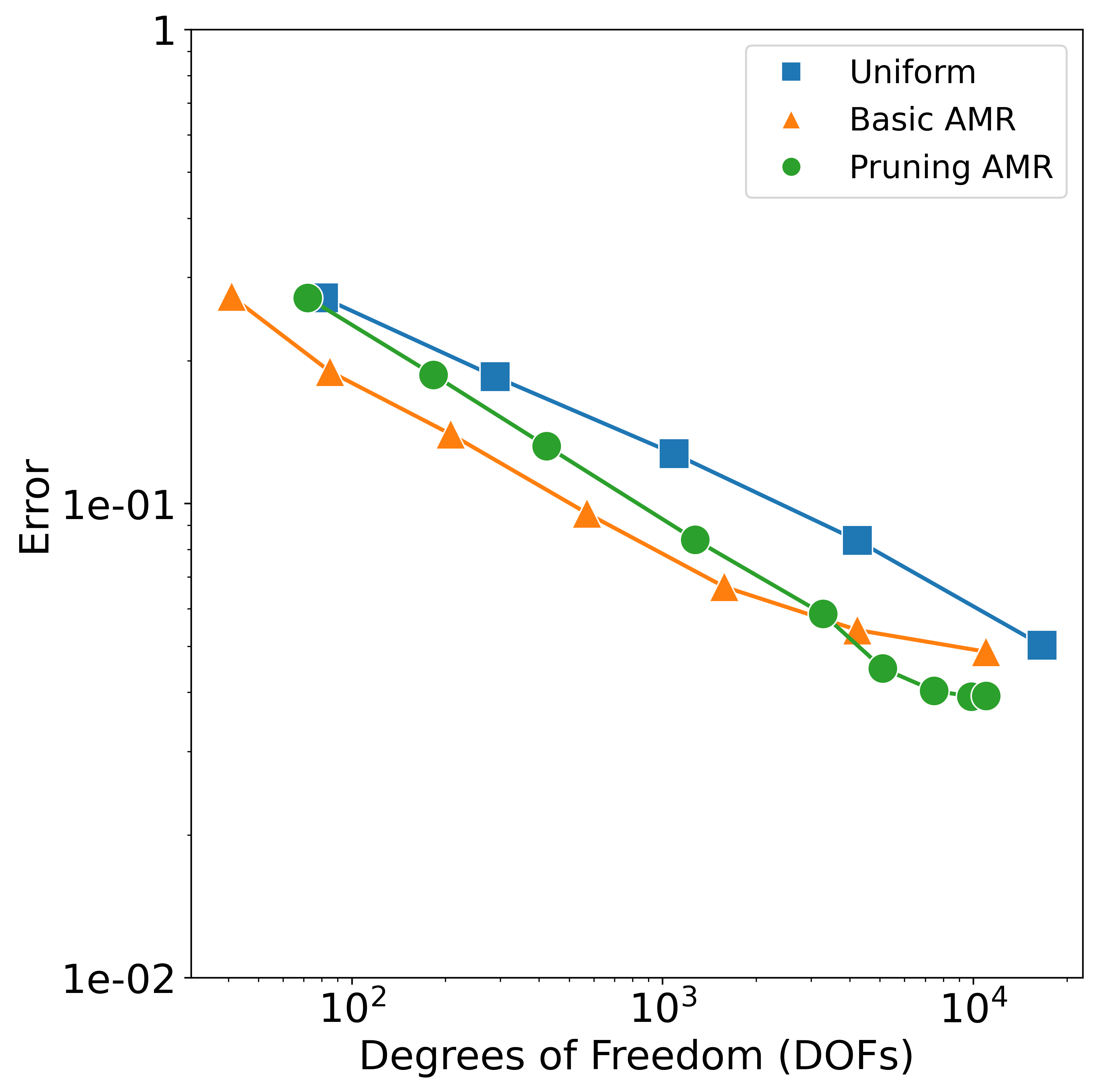}
    \end{minipage}%
    \hfill
    % Right: Table
    \begin{minipage}[c]{0.5\textwidth}
        \centering
		\text{At first iteration with $>10^4$ DOFs:}\\
		\text{}\\
        \begin{tabular}{|c|c|c|}
            \hline
				 & DOFs & error \\
            \hline
			Uniform 	& 16641 & 0.050 \\
			Basic AMR	& 10965	& 0.048 \\
			Pruning AMR	& 10962	& 0.039 \\
            \hline
        \end{tabular}
    \end{minipage}
    \caption{\textbf{Left:} Total error versus number of degrees of freedom plot is shown for \texttt{Uniform}, and the best-tuned instances of \texttt{BasicAMR} ($\tau=0.1$) and \texttt{PruningAMR} ($T=0.1$, $P=0.09$, and $\varepsilon = 10^{-3}$).  The \texttt{PruningAMR} method---i.e. Algorithm~\ref{alg:amr4inr}---drives down error at a faster rate than the \texttt{Uniform} approach and terminates with a lower error for an equivalent number of DOFs than either \texttt{BasicAMR} or \texttt{Uniform}. \textbf{Right:} Table of DOFs and error values for each algorithm at their final iteration.}
    \label{fig:err_dofs_2d}
\end{figure}

\begin{figure}[h]
\centering
\begin{tabular}{ccc}
\includegraphics[width=0.3\linewidth]{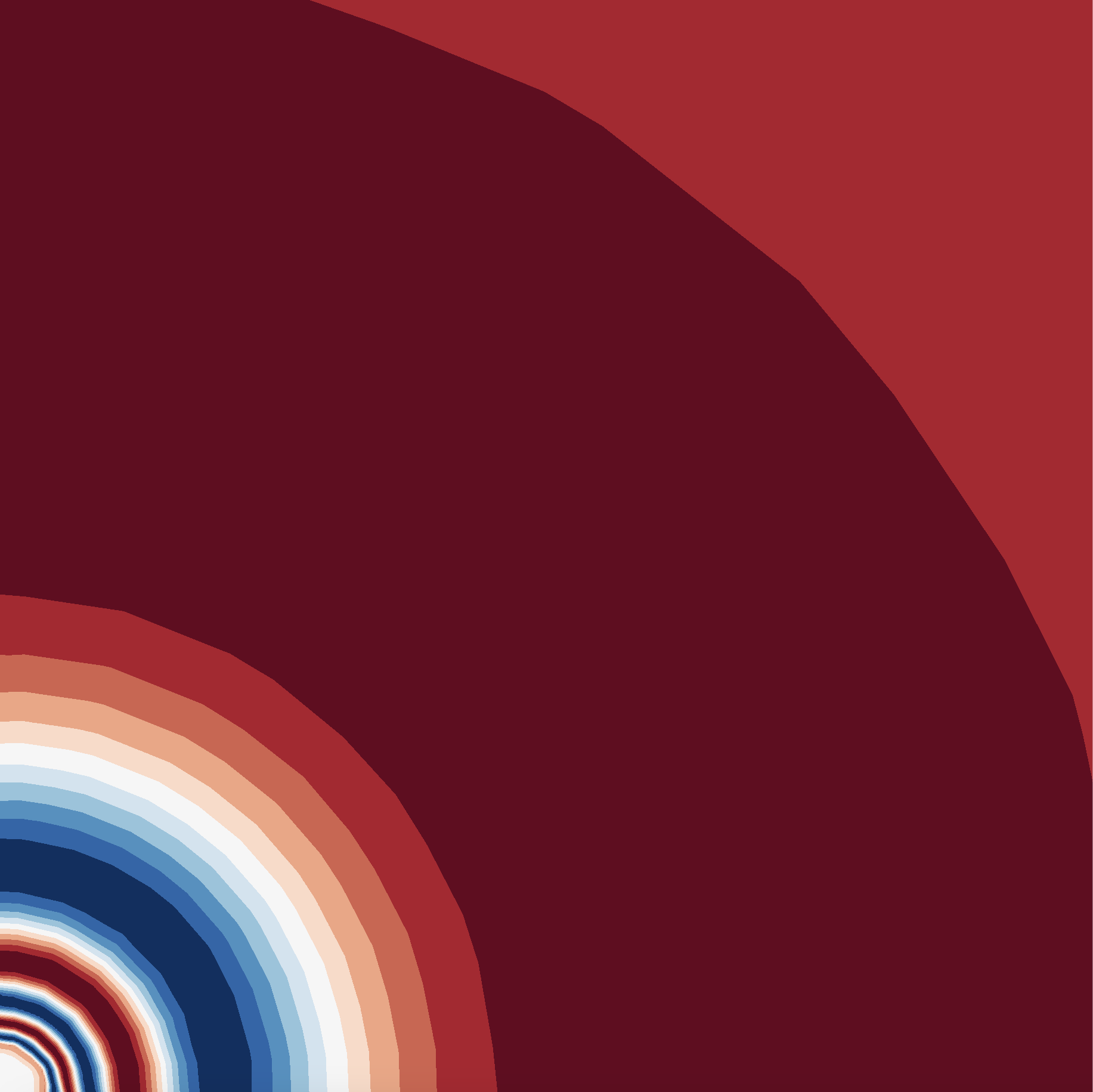}
& \includegraphics[width=0.3\linewidth]{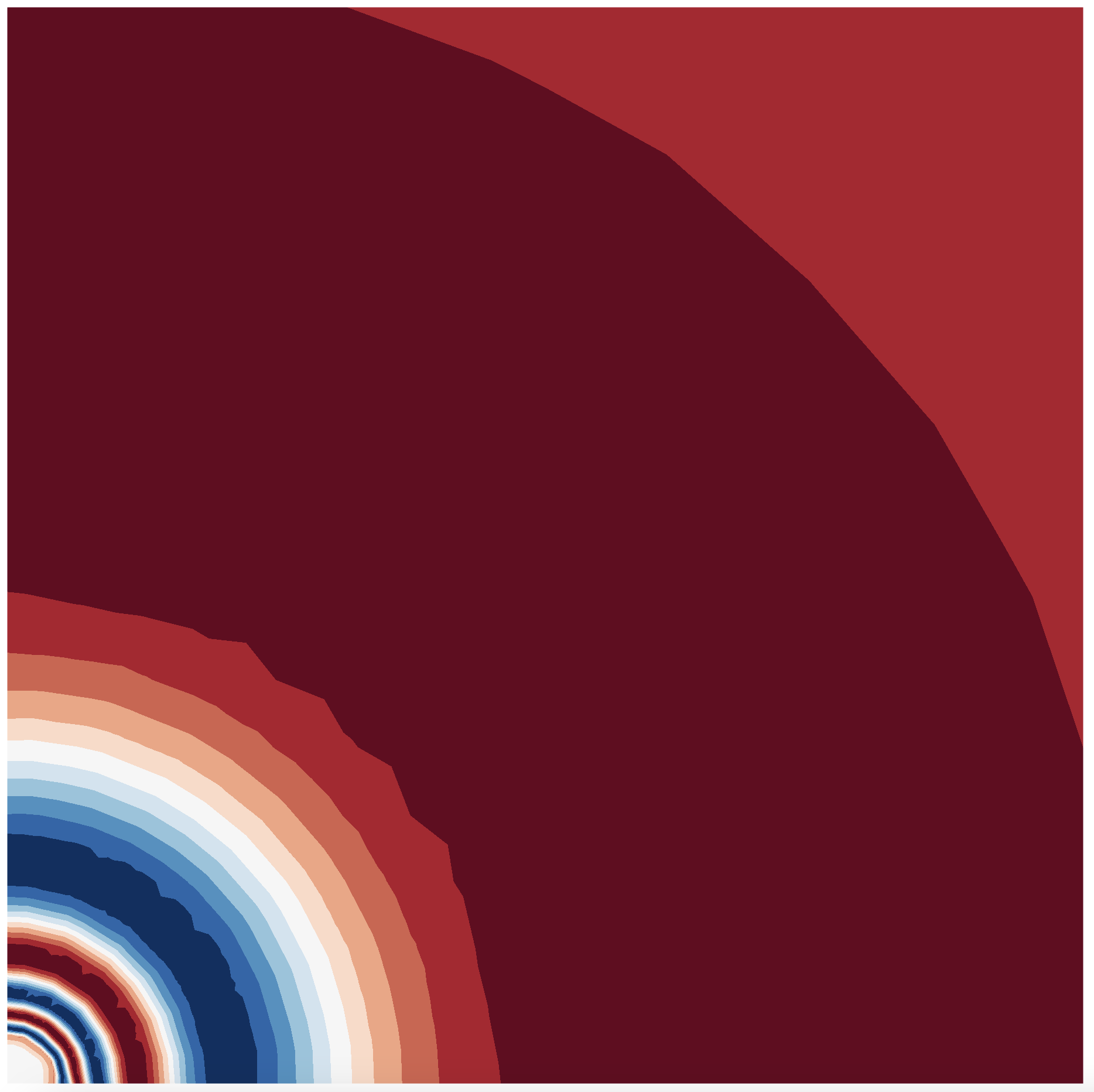}
&  \includegraphics[width=0.3\linewidth]{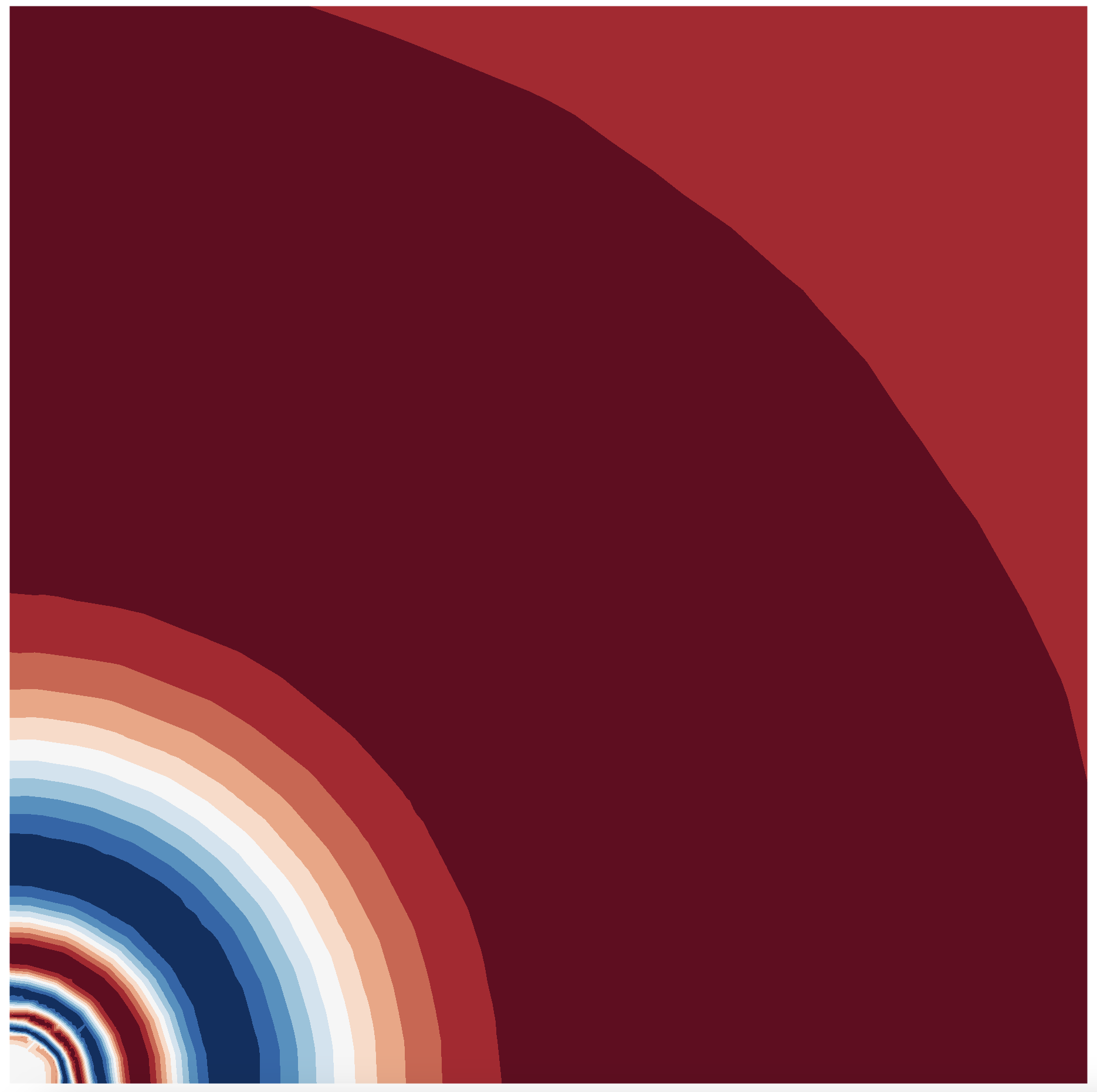} \\
\includegraphics[width=0.3\linewidth]{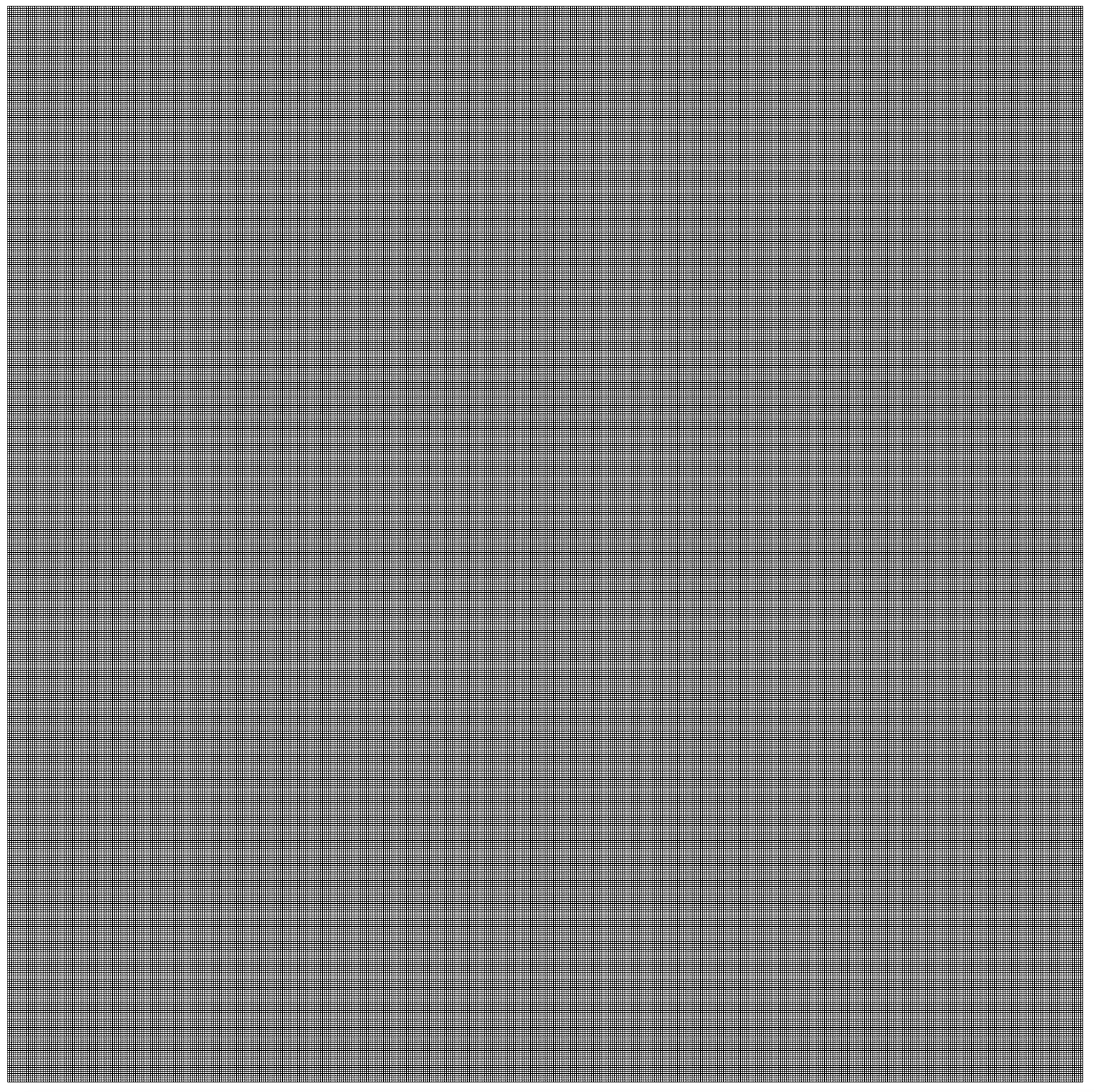}
& \includegraphics[width=0.3\linewidth]{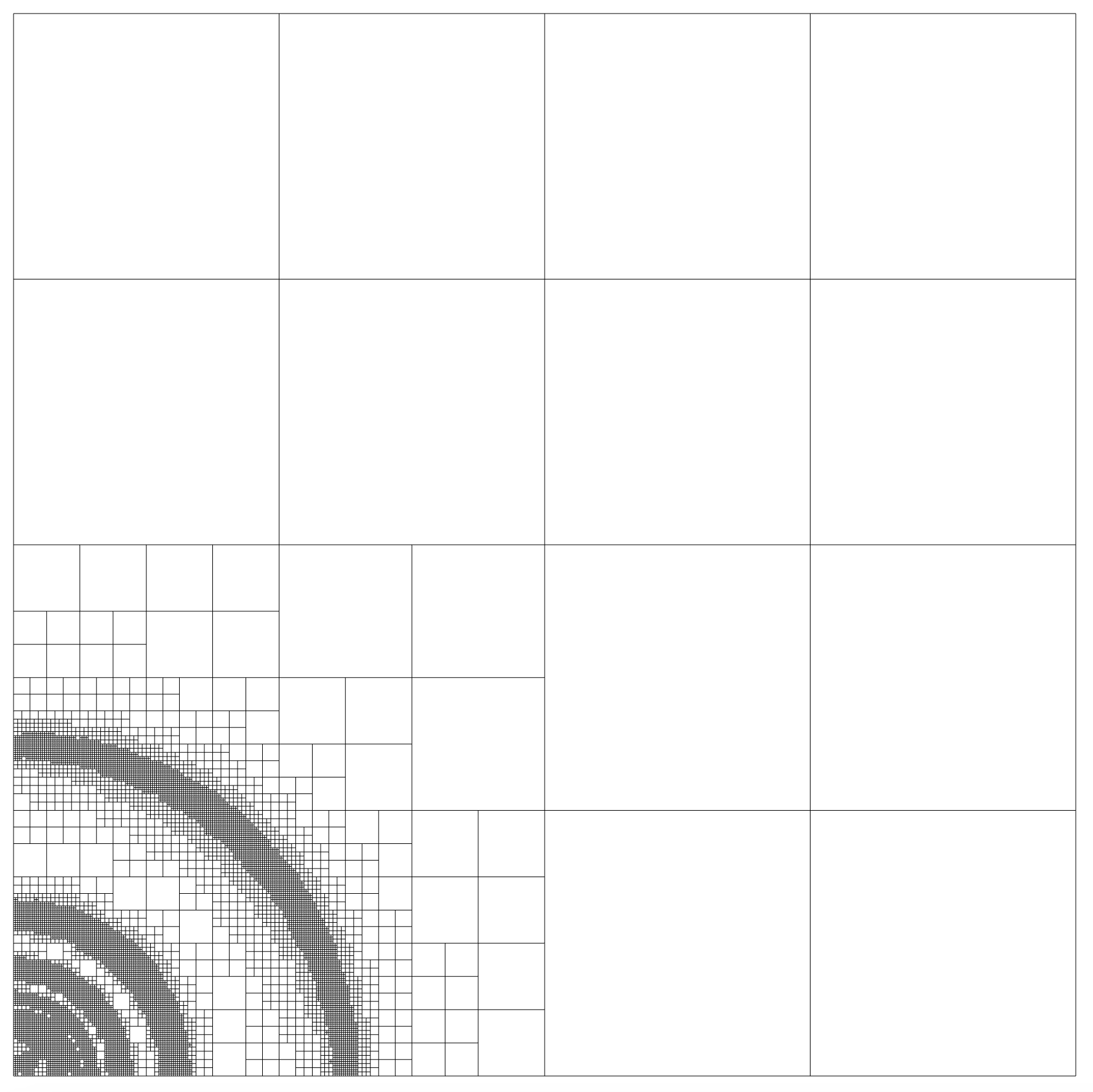}
&  \includegraphics[width=0.3\linewidth]{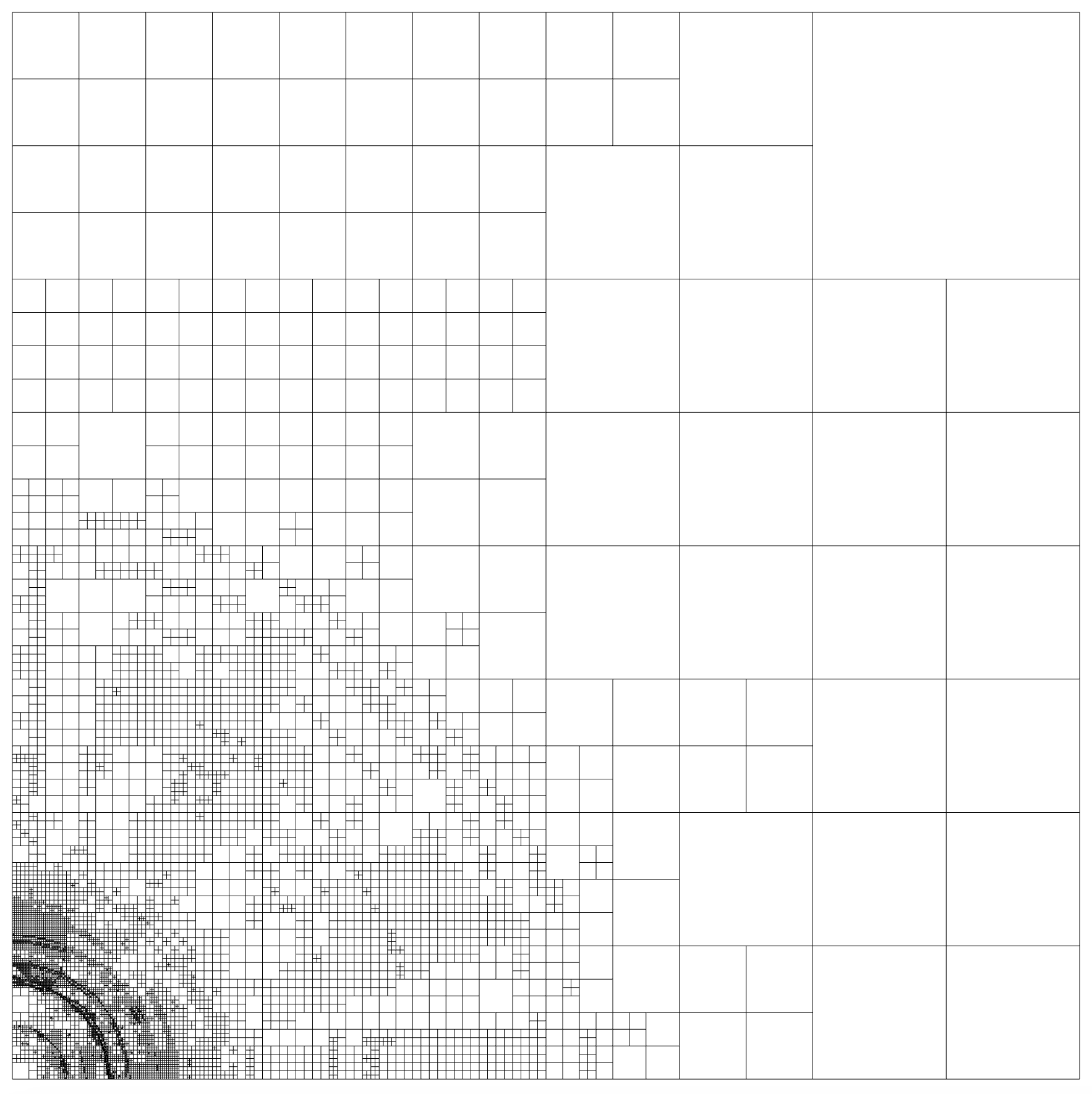} \\

\texttt{Uniform} & \texttt{BasicAMR} & \texttt{PruningAMR}
\end{tabular}
\caption{We compare three approaches to mesh refinement from a qualitative perspective for the 2D benchmark example.
The bottom row shows the mesh at the final state of the refinement method.
The top row shows a bilinear interpolant of the INR data evaluated at vertices of the mesh.
Treating \texttt{Uniform} as ``ground truth,'' observe that \texttt{BasicAMR} has multiple level sets with inaccurate variations, whereas \texttt{PruningAMR} is visibly more similar to the \texttt{Uniform} image.}
\label{fig:hos_viz}
\end{figure}

We verify and validate \texttt{PruningAMR} by testing on an INR fit to a benchmark function from the adaptive mesh refinement community~\cite[Section 2.8]{mitchell2013collection}.
Drawing samples of the function $f(r) := \sin(1/(\alpha + r))$ on $[0,1]^2$, where $r$ is the radius and $\alpha=1/50$, we train a simple ReLU feed-forward network with 4 layers of width 32. 
Sampled to the vertices of a regular mesh of $512\times512$ square elements, and visualized with bilinear interpolation, it is evident that the oscillations of the function have been captured to a fine resolution by the trained INR; see Figure~\ref{fig:hos_viz} (left column).
We use the open source software~\citet{mfem-web} to manage the adaptive meshing and~\citet{glvis-web} to generate the 2D figures.
For this example, we used 262,144 points to compute the errors shown in~\cref{fig:err_dofs_2d}.  % = (2^9)^dim with dim=2

We carried out experiments to study the effect of the key parameters for \texttt{PruningAMR} and \texttt{BasicAMR}, namely, $P$, $T$, and $\tau$.
The goal was to find parameters that minimize both total error and degrees of freedom at the termination of the algorithm. 
At a high level, the findings are consistent with what we expected.
If $P$, $T$, or $\tau$ are too low, too many elements are refined and the result is similar to that of \texttt{Uniform}.
If $P$ or $\tau$ is too high, too few elements are refined and \texttt{PruningAMR} stops after a few iterations.
For the 2D example with a maximum of 9 iterations and a dof threshold of 10,000, we found that for pruning we needed $P>0.05$ and $T>$1e-5, while for the basic method we needed 1e-3$<\tau<$0.2.
These choices of parameters are specific to the 2D example.

The best results for both \texttt{PruningAMR} and \texttt{BasicAMR} are shown in Figures~\ref{fig:err_dofs_2d} and~\ref{fig:hos_viz} .
The \texttt{Uniform} method drives error down linearly (in log scale) with respect to DOFs, as is expected.
The \texttt{BasicAMR} method (with optimal parameters) makes fewer refinements than \texttt{Uniform} in the first iteration, but drives down error at a similar rate to \texttt{Uniform}, until eventually leveling out.
The \texttt{PruningAMR} method (with optimal parameters)---i.e.~Algorithm~\ref{alg:amr4inr}---refines nearly all elements in the first iteration, but then drives down error at a \textit{faster} rate than \texttt{Uniform}, ultimately terminating at a lower error but equivalent DOF count as the \texttt{BasicAMR} method.
Furthermore, we show in Figure~\ref{fig:hos_viz} that the final mesh produced by~\texttt{PruningAMR} produces a qualitatively more accurate approximation to the INR than the final mesh produced by~\texttt{BasicAMR}.
We contend that this validates the effectiveness of Algorithm~\ref{alg:amr4inr} as a means for adaptive mesh refinement as the need to tune parameters is a challenge affecting all adaptive refinement schemes.

\subsection{Example 1: Physics-Informed Neural Network for Navier--Stokes}
% define what a PINN is
A physics-informed neural network, or PINN, is a neural network that approximates the solution to a differential equation~\citep{PINN}. Typically, the PINN is trained by encoding the differential equation in the loss function and assumes no access to solution data for training purposes. 
%why PINNs are INRs
PINNs are an example of an INR whose input is a coordinate in space or space-time and whose output is the approximate solution to a differential equation, possibly with multiple components. Notably, because PINNs are usually trained over randomized points in a domain, a discretization method is needed to visualize the approximated solution. 

In this example, we apply our algorithm to a neural network representing the solution to the incompressible Navier--Stokes equations for fluid flow past a circular cylinder in 2D space plus time. 
% give more info on example
% explain PINN architecture and domain
We use a network with 8 hidden layers, 20 neurons per layer, and the hyperbolic tangent activation function $\tanh(x)$. 
The input to the PINN is $x \in [1,5]$, $y\in [-2,2]$, and $t\in[0,4]$.
The output of the PINN contains three components: the velocity in the x-direction and y-direction, and pressure. We will focus on visualizing the velocity in the x-direction, $u$. However, the same algorithm can be applied for either of the other PINN outputs. 
This PINN can be trained using the algorithm and corresponding repository presented in~\citet{NSPINN}. 

All results for this example use the hyperparameters $T=\tau=10^{-3}$, $P=0.15$, $\varepsilon = 0.005$, $K_{max} = 5$, and $n_{ID} = 256$. We use $n_{err}=256$ for \texttt{PruningAMR} and $n_{err} = 512$ for \texttt{BasicAMR}. We use $2097152$ randomly sampled points to compute the root mean squared error for all methods. All visualizations for this example are created using ParaView~\citep{Paraview}.

\begin{figure}[ht]
\centering
\begin{tabular}{ccc}
\includegraphics[width=0.25\linewidth]{"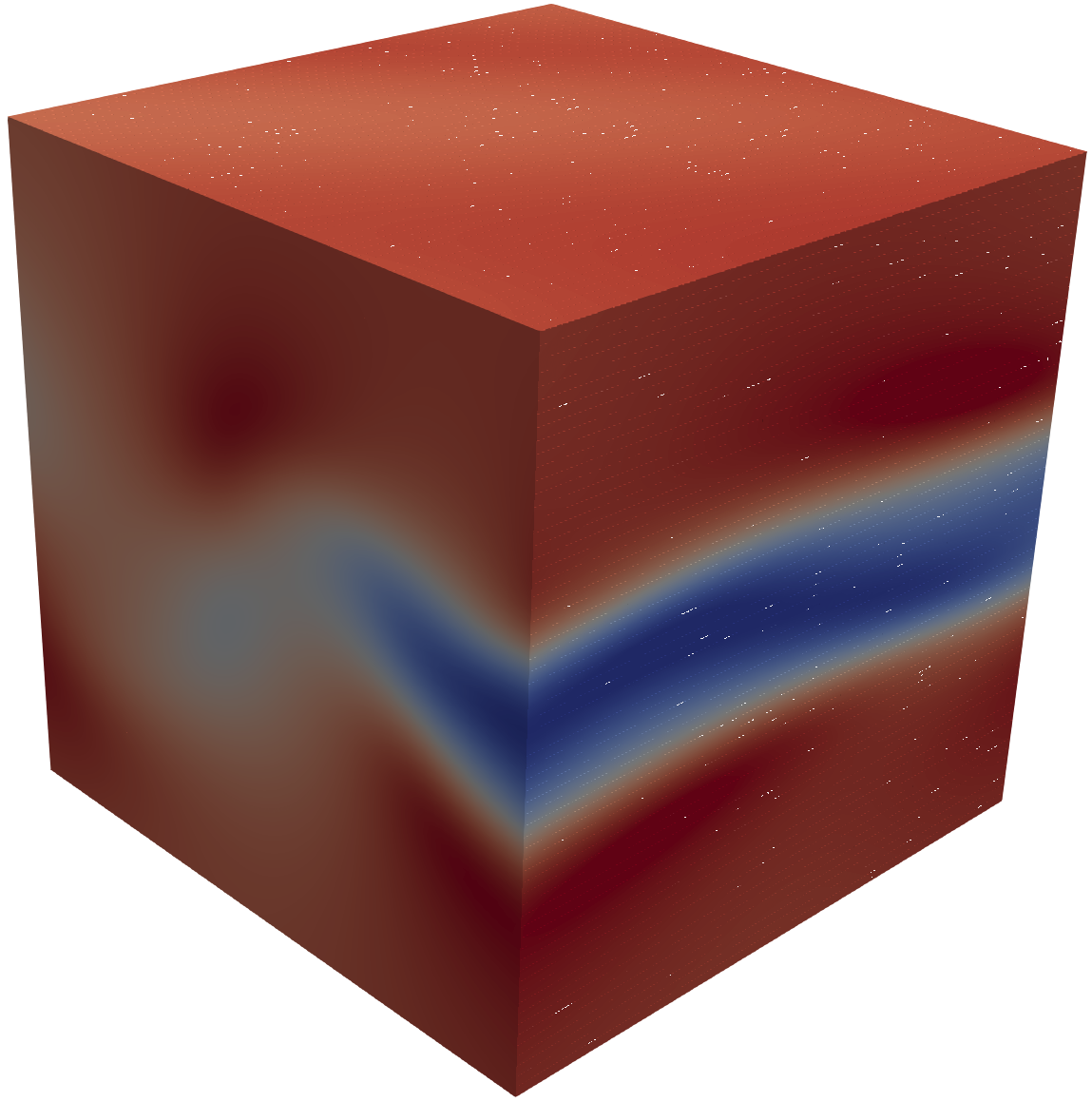"}
&\hspace{0.9cm}  \includegraphics[width=0.25\linewidth]{"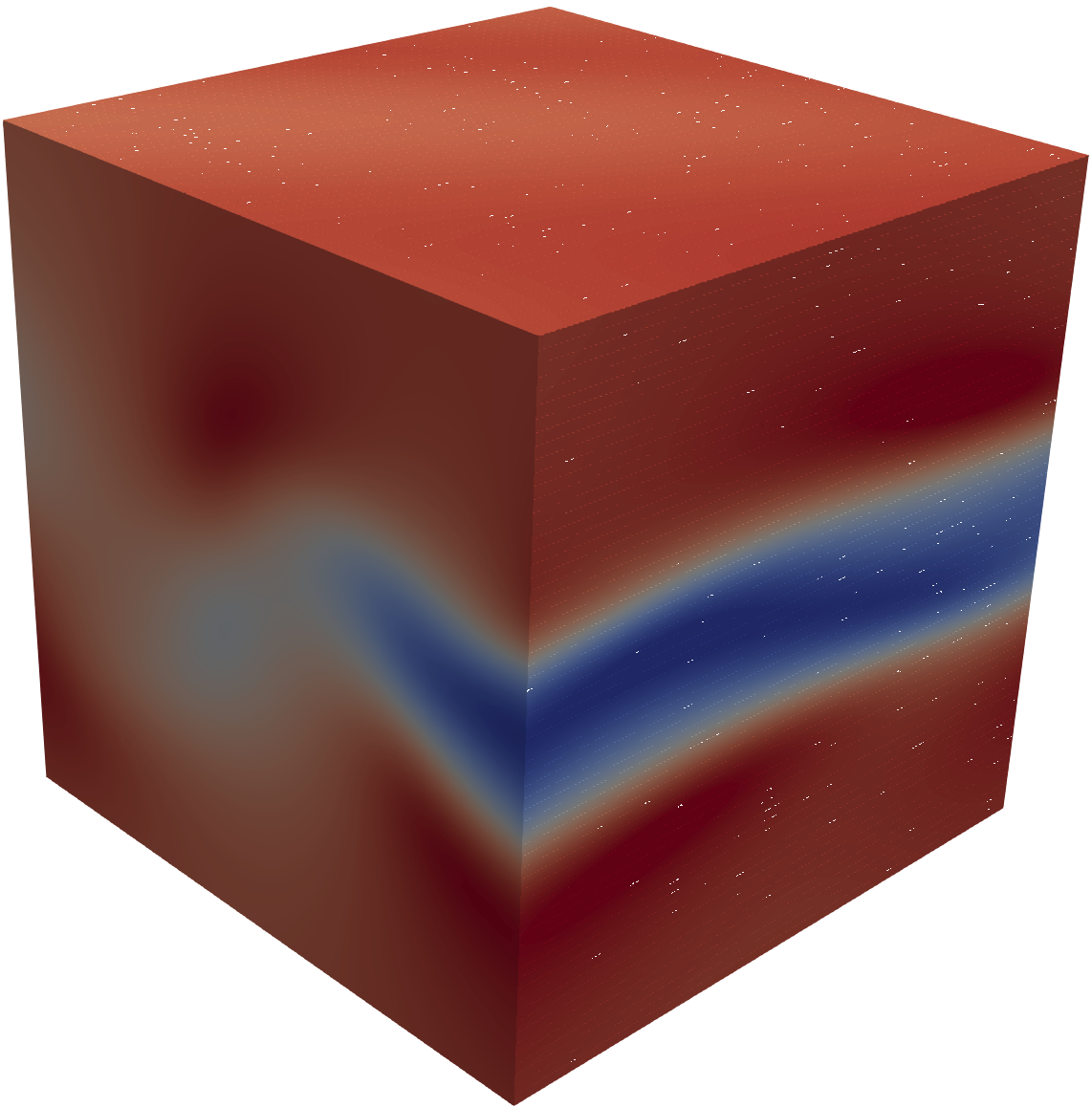"} \hspace{0.9cm}
&  \includegraphics[width=0.25\linewidth]{"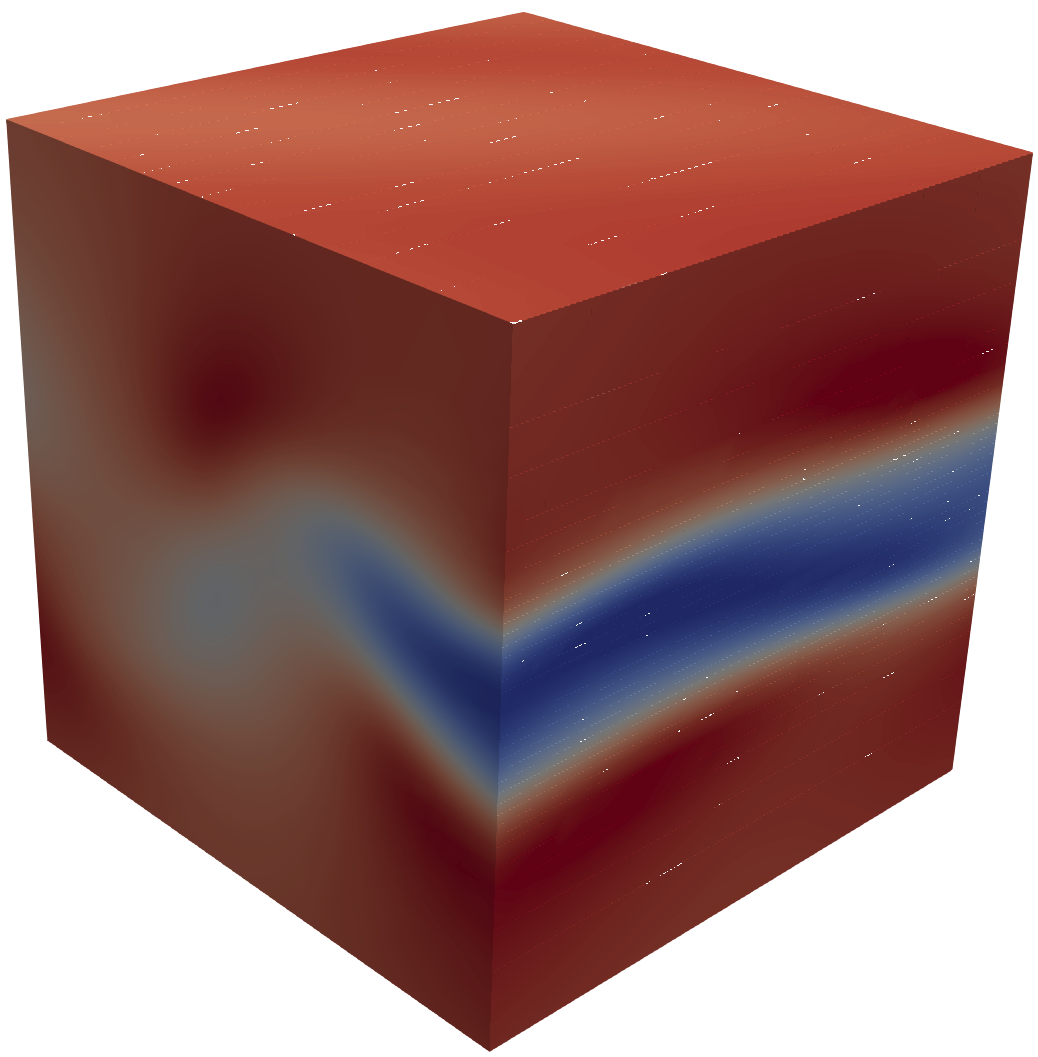"} \\

\texttt{Uniform} & \texttt{BasicAMR} & \texttt{PruningAMR}
\end{tabular}
\caption{Comparison of 3D visualizations of the Navier--Stokes PINN created by \texttt{Uniform},  \texttt{BasicAMR} ($\tau = 10^{-3}$), and \texttt{PruningAMR} ($T=10^{-3}$, $P=0.15$, $\varepsilon = 0.005$) refinement for fluid flow past a circular cylinder. Note that two dimensions are in space and one is in time. The difference between each visualization appears negligible.}
\label{fig:PINN_3D}
\end{figure}

\begin{figure}[ht]
\centering
    % Left: Image
    \begin{minipage}[c]{0.45\textwidth}
        \centering
        \includegraphics[width=\linewidth]{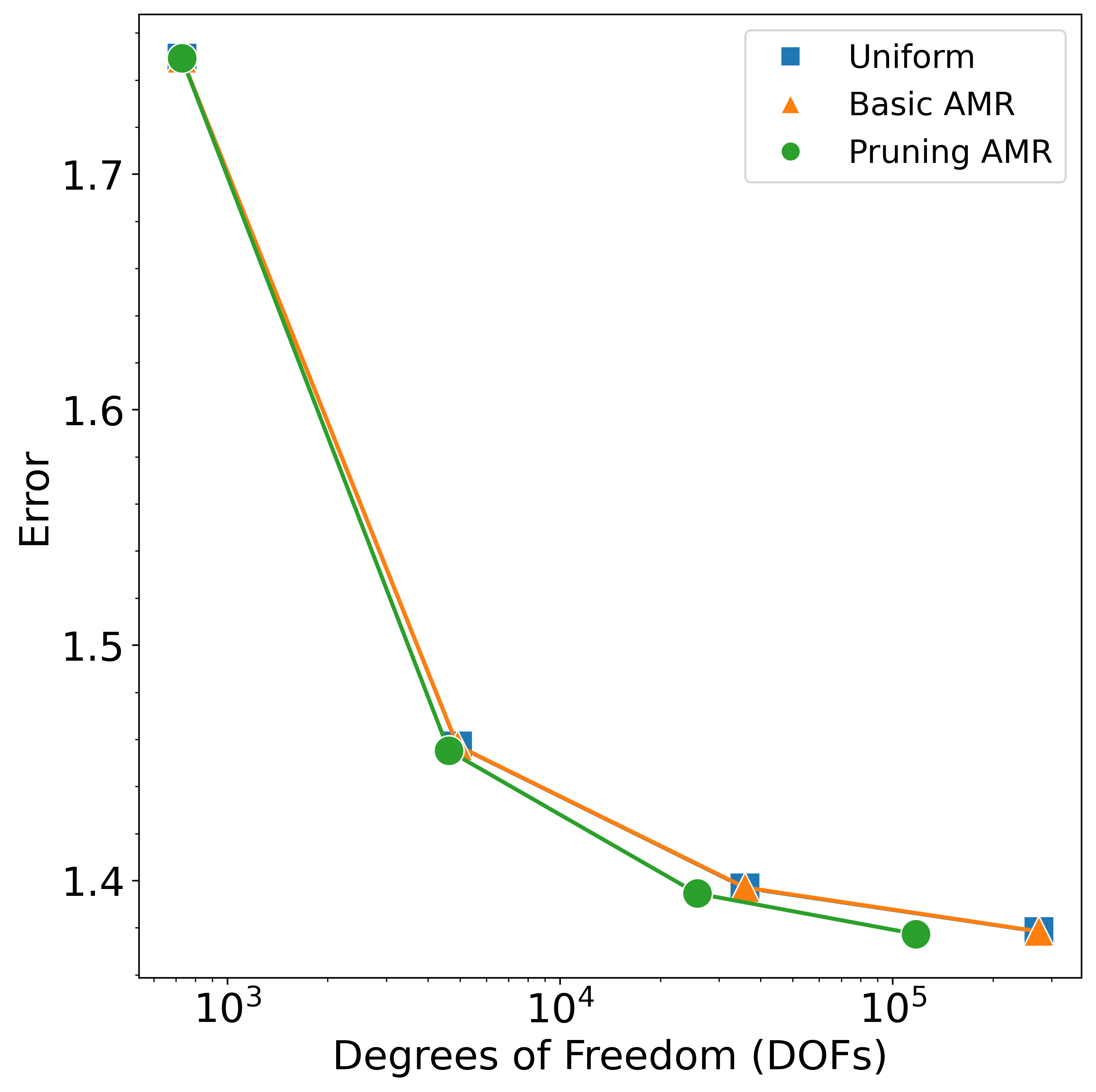}
    \end{minipage}%
    \hfill
    % Right: Table
    \begin{minipage}[c]{0.5\textwidth}
        \centering
    \text{At fifth iteration:}\\
    \text{}\\
        \begin{tabular}{|c|c|c|}
            \hline
         & DOFs & error \\
            \hline
      Uniform   & 274625 & 1.378 \\
      Basic AMR & 274625 & 1.378 \\
      Pruning AMR & 117225 & 1.377 \\
            \hline
        \end{tabular}
    \end{minipage}
\caption{\textbf{Left}: Total error versus number of degrees of freedom plot is shown for \texttt{Uniform},  \texttt{BasicAMR} ($\tau=10^{-3}$), and \texttt{PruningAMR} ($T=10^{-3}$, $P=0.15$, $\varepsilon = 0.005$) for visualizing the Navier--Stokes PINN.  \texttt{PruningAMR} achieves a lower number of DOFs for a similar error to \texttt{BasicAMR} and \texttt{Uniform}. \textbf{Right:} Table of DOFs and error values for each algorithm at their final iteration.}
\label{fig:PINN_plot}
\end{figure}

A 3D (2D space + time) representation visualization of the PINN for each of the three algorithms is shown in Figure~\ref{fig:PINN_3D}. Visually, each representation looks the same. However, the corresponding DOFs vs error plot, Figure~\ref{fig:PINN_plot}, indicates that the final \texttt{PruningAMR} mesh uses 57\% fewer DOFs than the final \texttt{BasicAMR} and \texttt{Uniform} meshes.
This provides a non-trivial example of the practical efficiency of our approach.
We also observe that the \texttt{BasicAMR} results in uniform refinement at each iteration, rendering it ineffective as a driver of effective AMR for this example.

% error vs dofs plot

\begin{figure}[H]
\centering
\begin{tabular}{cccc}
$t=0$  & \includegraphics[width=0.25\linewidth]{"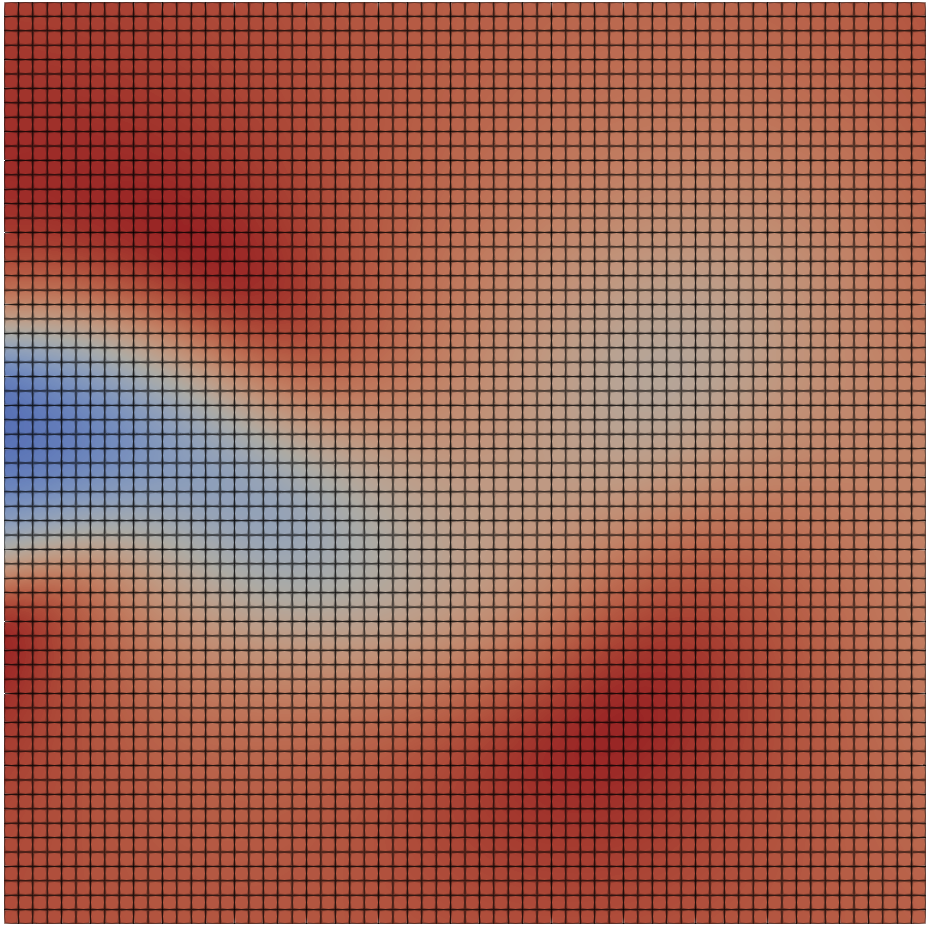"}
& \includegraphics[width=0.25\linewidth]{"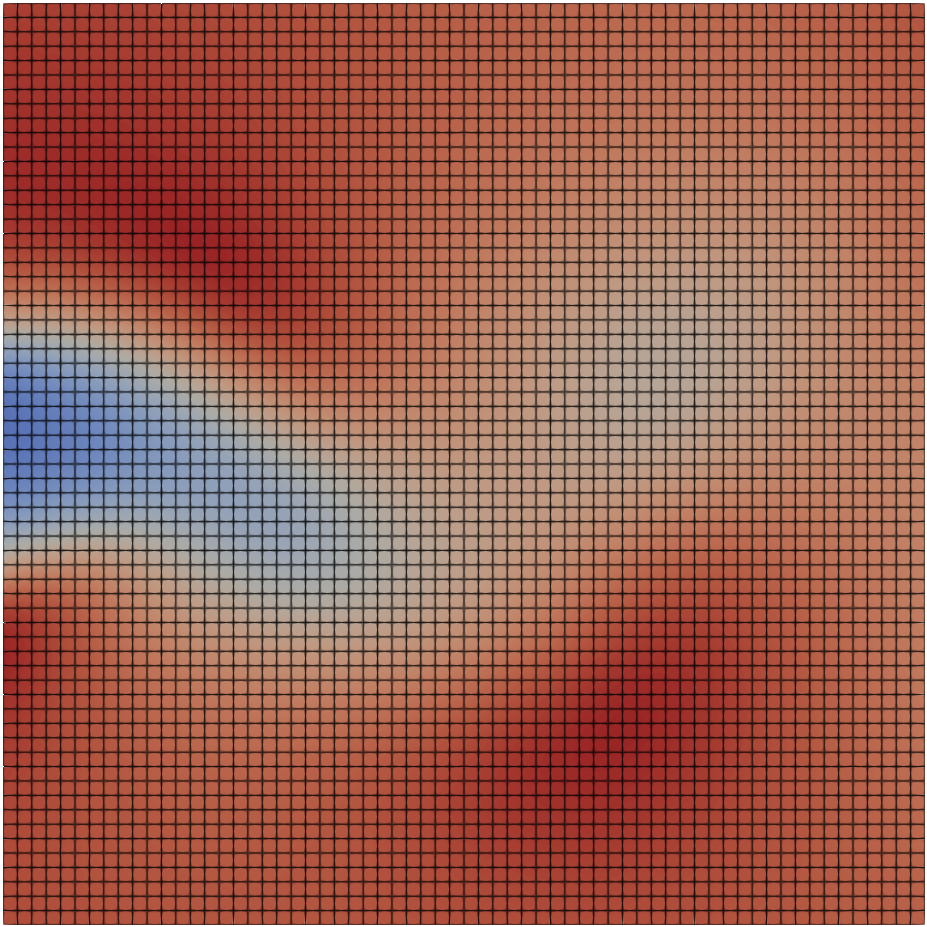"}
&  \includegraphics[width=0.25\linewidth]{"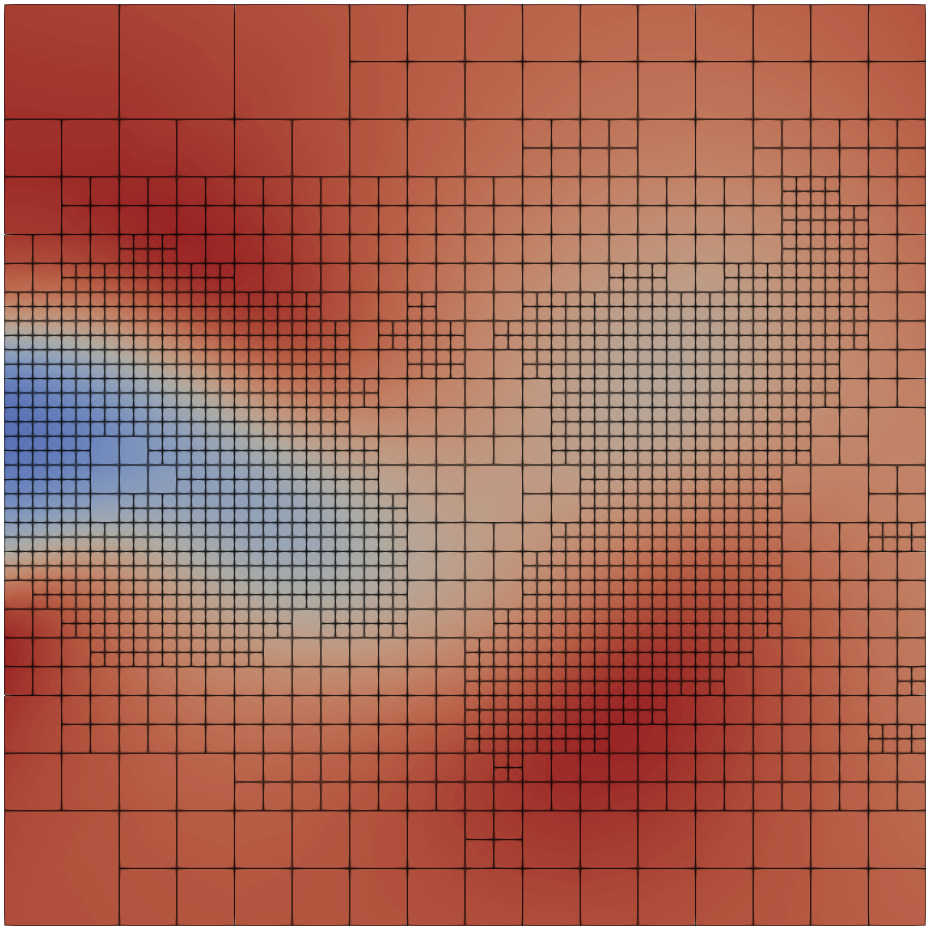"} \\
$t=2$ & \includegraphics[width=0.25\linewidth]{"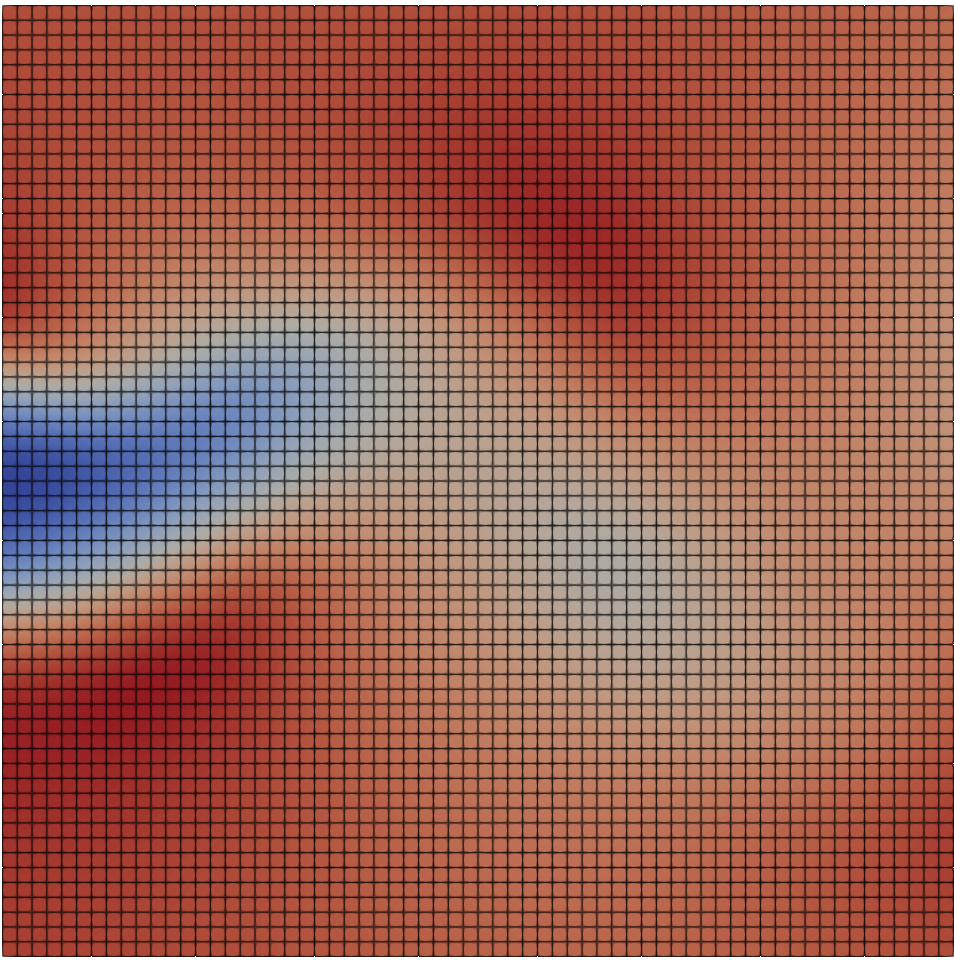"}
& \includegraphics[width=0.25\linewidth]{"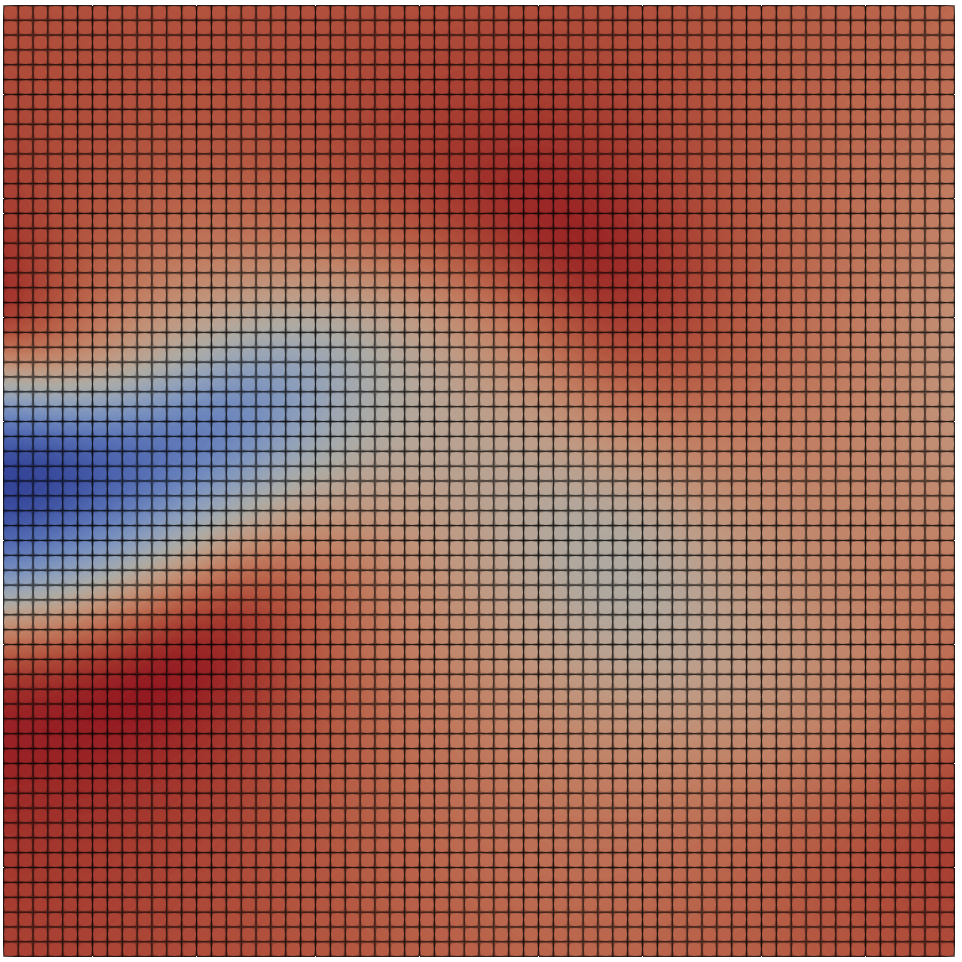"}
&  \includegraphics[width=0.25\linewidth]{"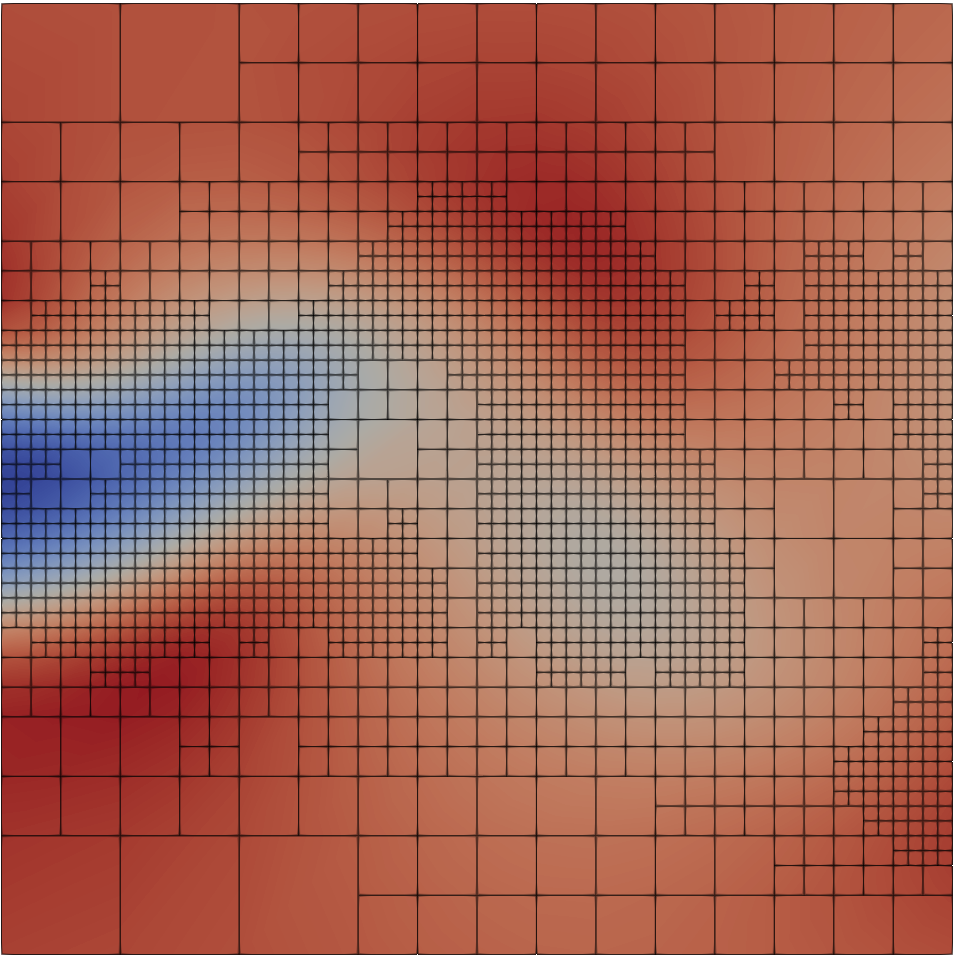"} \\
$t=4$ & \includegraphics[width=0.25\linewidth]{"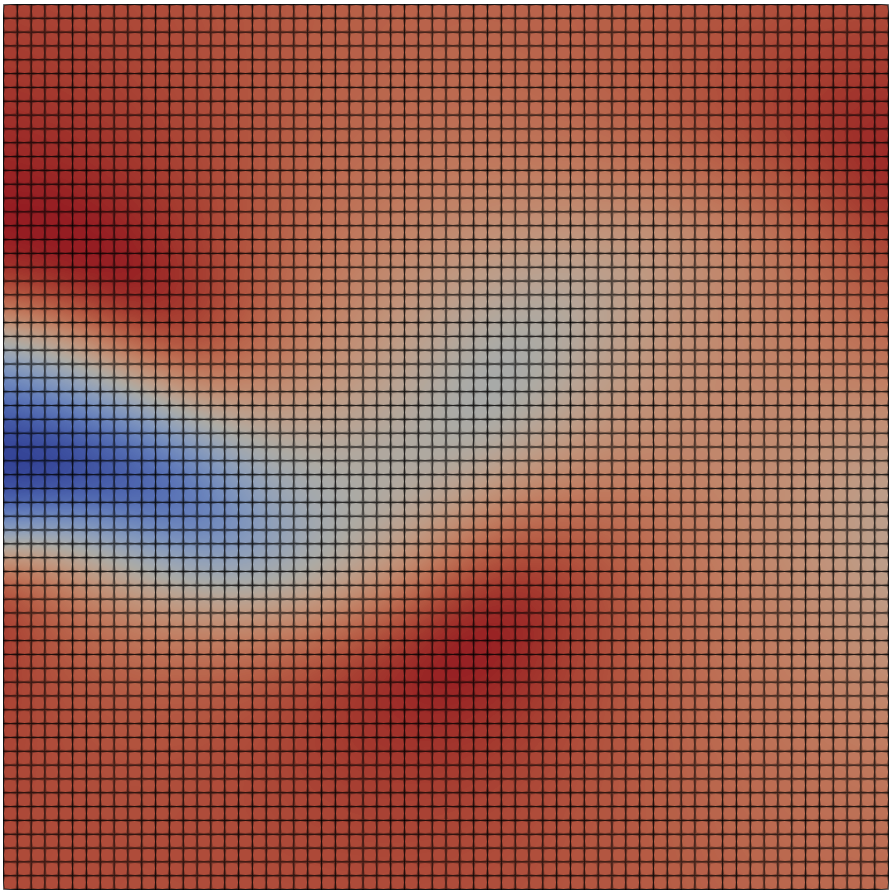"}
& \includegraphics[width=0.25\linewidth]{"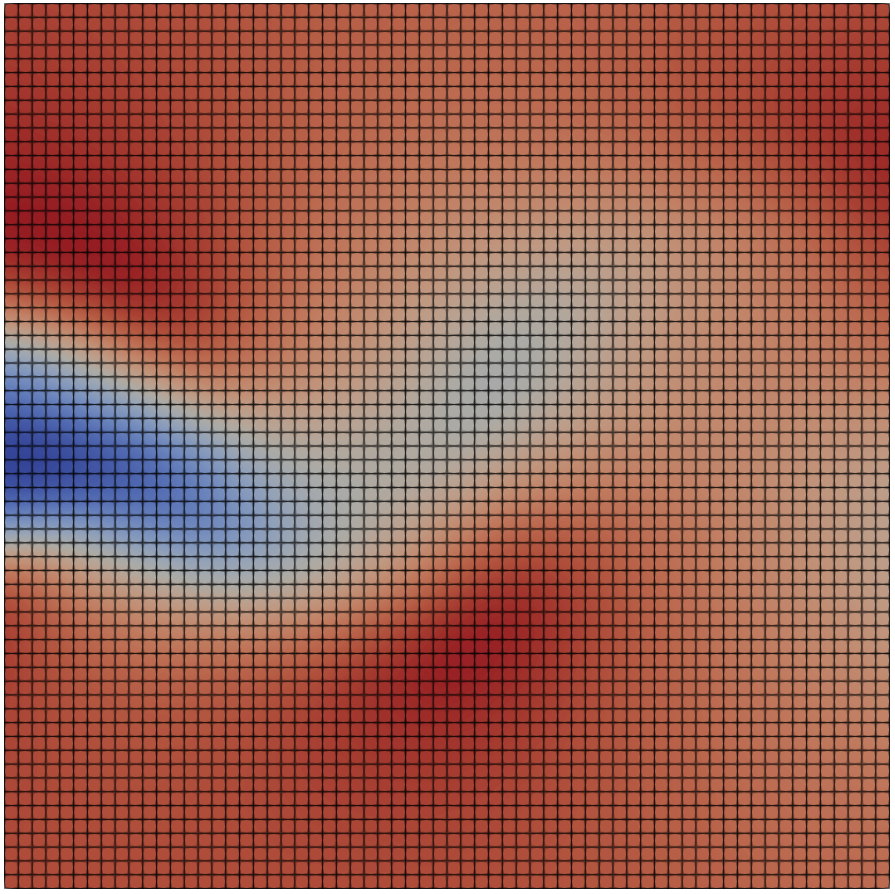"}
&  \includegraphics[width=0.25\linewidth]{"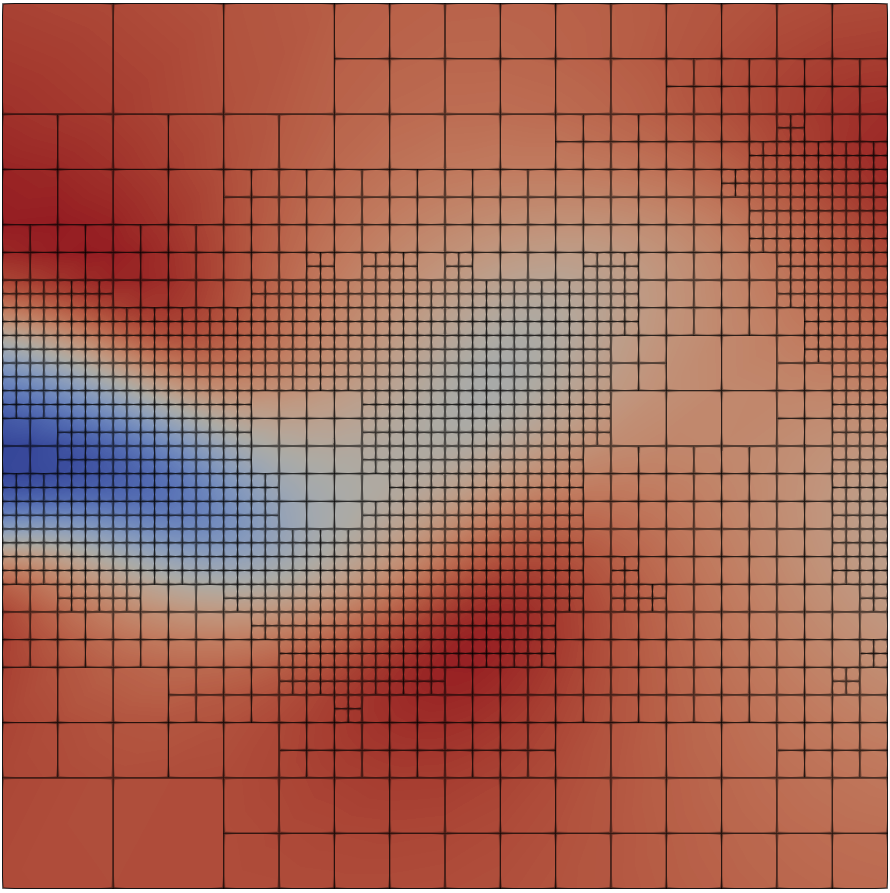"} \\

& \texttt{Uniform} & \texttt{BasicAMR} & \texttt{PruningAMR}
\end{tabular}
\caption{Comparison between meshes created by \texttt{Uniform},  \texttt{BasicAMR} ($\tau = 10^{-3}$), and \texttt{PruningAMR} ($T=10^{-3}$, $P=0.15$, $\varepsilon = 0.005$) refinement for the Navier--Stokes PINN. Top row: t=0, middle row: t=2, bottom row: t=4. Each figure shows the result of five iterations of refinement. For each row, the images are visually similar, but \texttt{PruningAMR} uses fewer elements than the other two methods. \texttt{PruningAMR} concentrates its elements on the wake from the cylinder, and uses fewer elements in the surrounding regions.}
\label{fig:PINN_time_slices}
\end{figure}

We note that for this example, we have treated the domain as a 3D mesh, with no special consideration given to the time dimension.
This is atypical when designing adaptive meshes for traditional finite elements, since numerical schemes usually require a uniform time step across the spatial domain.
If such a mesh is required, the user does can specify that the algorithm be applied to a pre-defined collection of (2D) time slices instead. 
Either way, the resulting mesh can be sliced at a requested time and visualized for inspection.
To demonstrate this ability and some of the resulting adaptive meshes, time slices for \texttt{Uniform}, \texttt{BasicAMR}, and \texttt{PruningAMR} are shown in Figure~\ref{fig:PINN_time_slices}.

\subsection{Example 2: Simulated dynamic CT INR} \label{sec:simulation_example}
\begin{figure}
\centering
  \begin{tabular}{ccc}
\includegraphics[width=0.3\linewidth]{"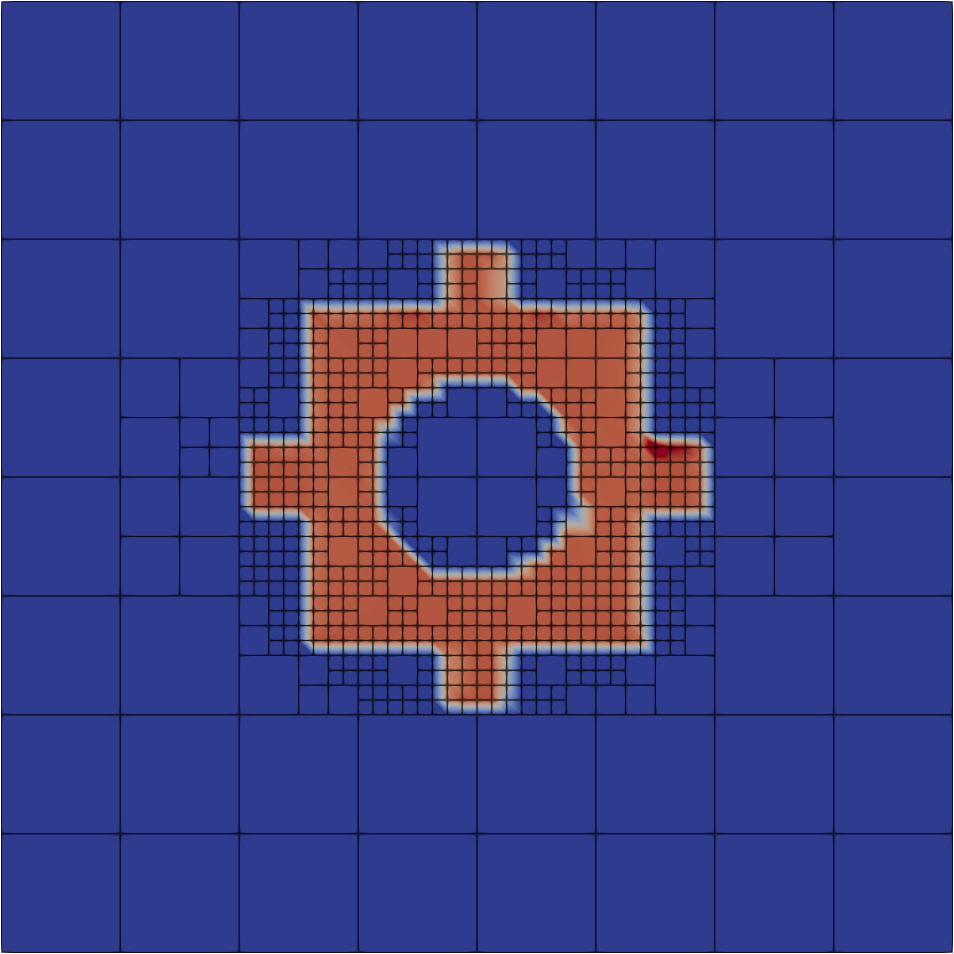"}
& \includegraphics[width=0.3\linewidth]{"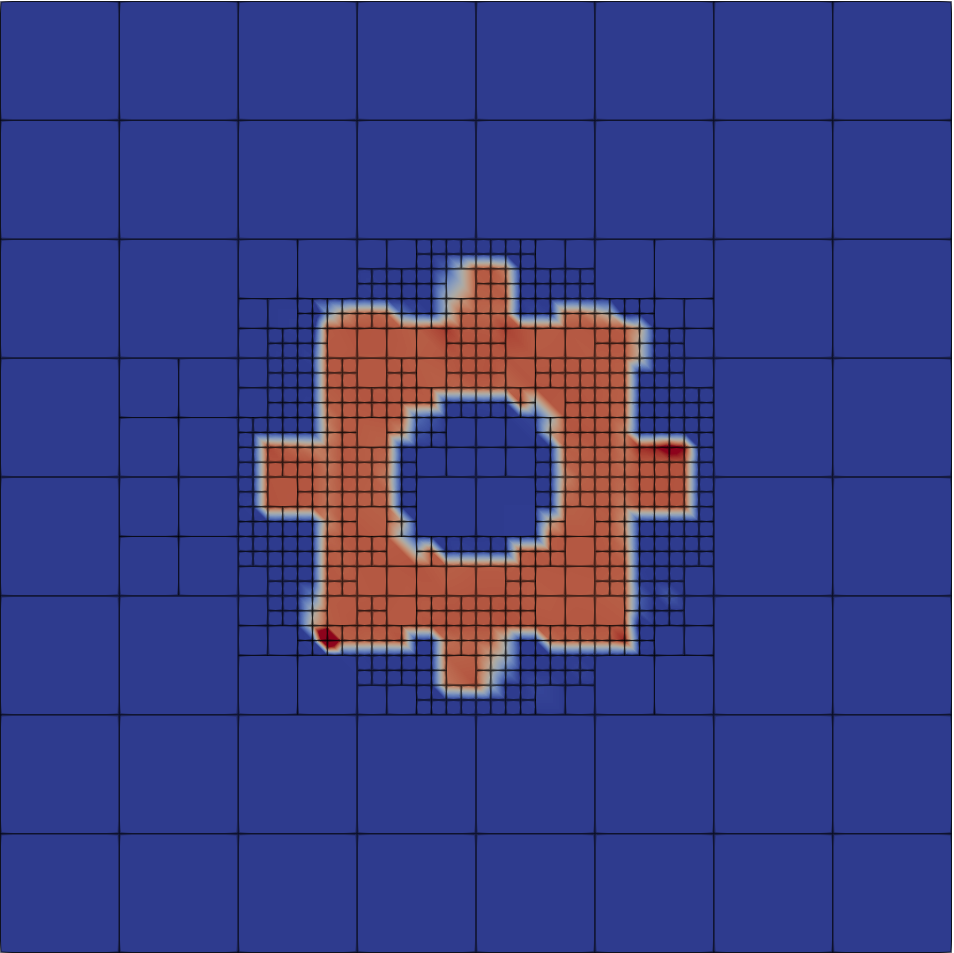"}
&  \includegraphics[width=0.3\linewidth]{"./CT5_pruned_t=1_yslice_new.png"} \\
$t = -1$ & $t = 0$ & $t = 1$
\end{tabular}
  \caption{Multiple time slices of simulated CT INR visualized using \texttt{PruningAMR} ($T=10^{-4}$, $P=0.075$, $\varepsilon = 10^{-3}$) AMR. Notice that the mesh changes with time as the object changes shape.}
  \label{fig:CT5_time_slices}
\end{figure}

We now consider an INR from a simulated CT scan of a 3D object being compressed in time. 
The object is a cube with a cylindrical hole missing from its center. 
At time $t= - 1$ the cube is uncompressed, but over time the cube is compressed on four sides by rectangular prisms. 
See the leftmost image in Figure~\ref{fig:CT5_time_slices} for an overhead view. 
More information about the pre-trained INR can be found in~\citet{CT_INR}. 
The architecture of the INR consists of a Gaussian random Fourier feature encoding layer (see~\citet{tancik2020fourier}), five fully-connected layers, each with a width of 256 neurons, swish activation functions, and a linear output layer with scalar output. The inputs to the INR are $x,y,z,t$, each in the range $[-1,1]$. 
All visualizations for this example are created using ParaView~\citep{Paraview}.

\begin{figure}[h]
    \centering
    % Left: Image
    \begin{minipage}[c]{0.45\textwidth}
        \centering
        \includegraphics[width=\linewidth]{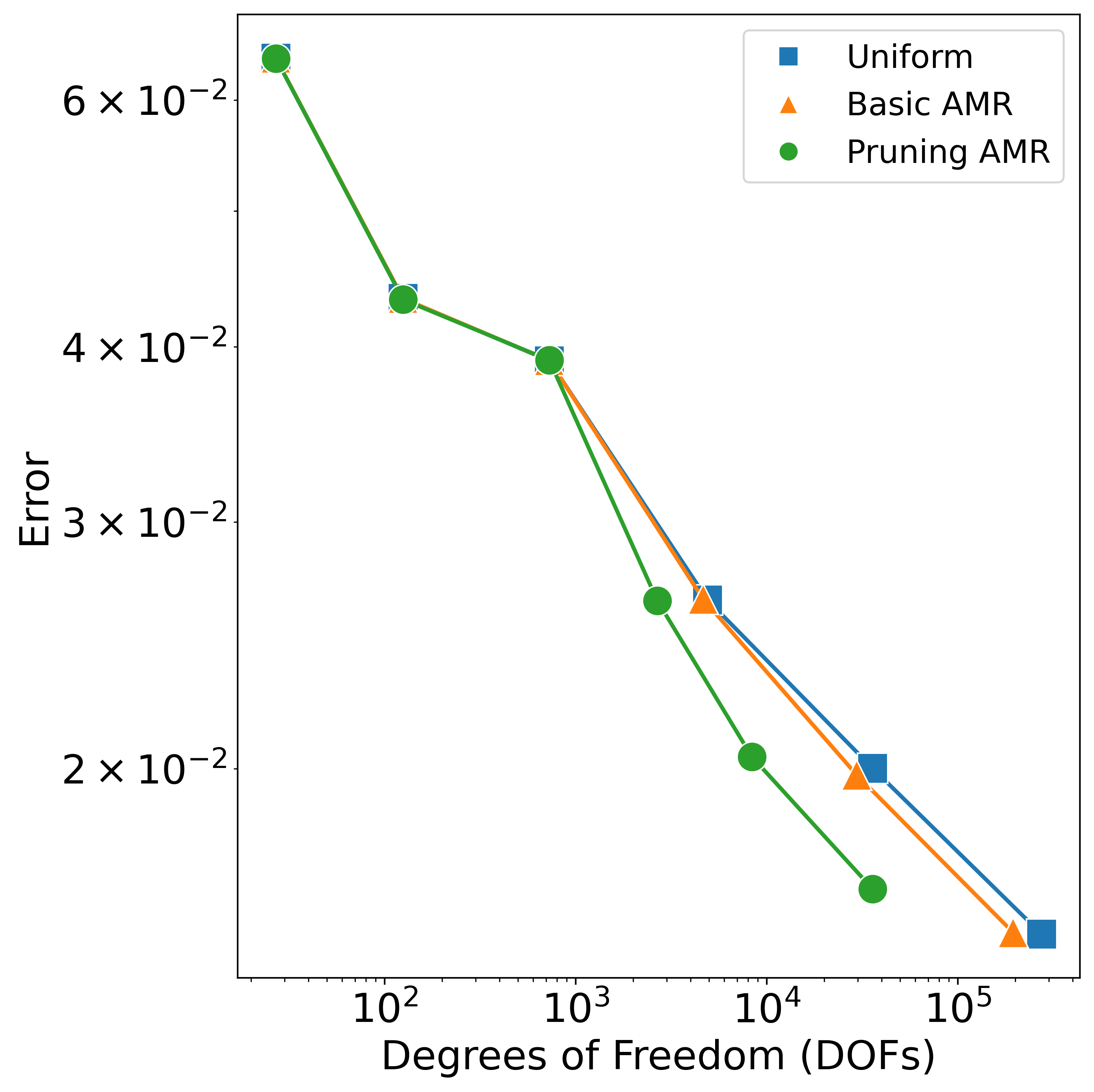}
    \end{minipage}%
    \hfill
    % Right: Table
    \begin{minipage}[c]{0.5\textwidth}
        \centering
    \text{At sixth iteration:}\\
    \text{}\\
        \begin{tabular}{|c|c|c|}
            \hline
         & DOFs & error \\
            \hline
      Uniform   & 274625 & 0.0152 \\
      Basic AMR & 193242 & 0.0153\\
      Pruning AMR & 35994 & 0.0167 \\
            \hline
        \end{tabular}
    \end{minipage}

    \caption{\textbf{Left:} Total error versus number of degrees of freedom are shown for \texttt{Uniform}, \texttt{BasicAMR} ($\tau = 10^{-4}$), and \texttt{PruningAMR} ($T=10^{-4}$, $P=0.075$, $\varepsilon = 10^{-3}$) visualization of a simulated CT scan of a 3D object being compressed over time. \texttt{PruningAMR} achieves the same accuracy as the other two methods with significantly fewer DOFs. The gap in DOFs increases with each iteration. \textbf{Right:} Table of DOFs and error values for each algorithm at their final iteration.}
    \label{fig:compare_CT5}
\end{figure}

We applied \texttt{PruningAMR} to the simulated CT INR and compared it to \texttt{Uniform} refinement and \texttt{BasicAMR}, as described in Section~\ref{sec:2D_example}. 
All results for this example use the hyperparameters: $T = \tau = 10^{-4}$, $P=0.075$, $\varepsilon=10^{-3}$, $K_{max} = 6$, and $n_{ID} = 256$. 
We use $n_{err} = 256$ for \texttt{PruningAMR} and $n_{err} = 512$ for \texttt{BasicAMR}.
We found these hyperparameters empirically, by keeping $n_{err}$ and $\varepsilon$ fixed and varying the accuracy thresholds ($T, \tau, P$) to target maximal accuracy within five iterations. 
We use $1048576$ randomly sampled points to compute the root mean squared error for all methods. 

\begin{figure}
\centering
\begin{tabular}{ccc}
\includegraphics[width=0.3\linewidth]{"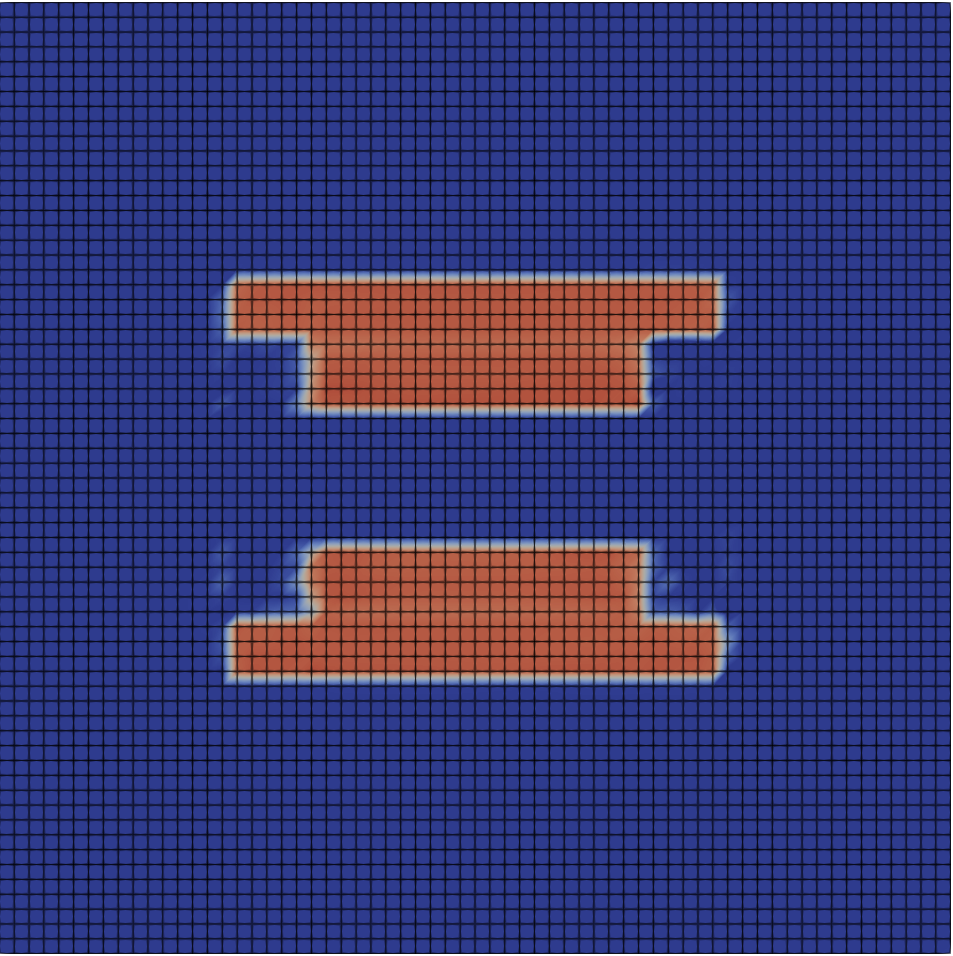"}
& \includegraphics[width=0.3\linewidth]{"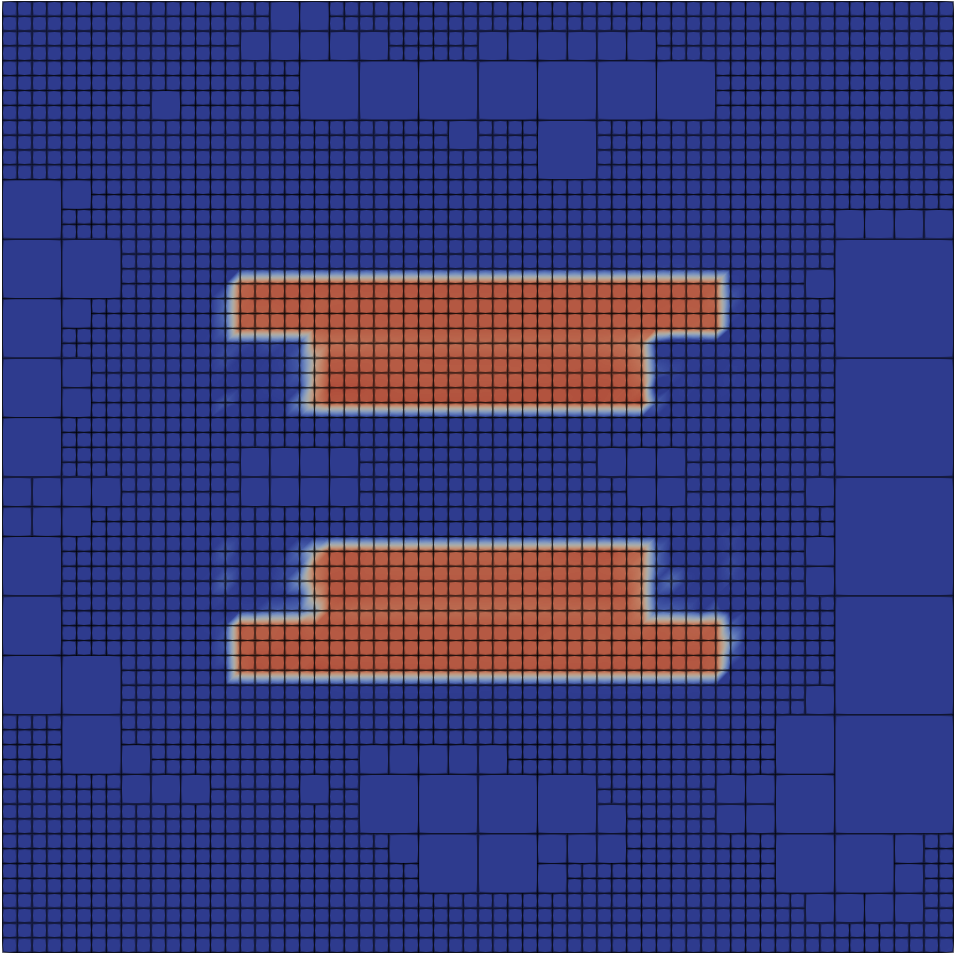"}
&  \includegraphics[width=0.3\linewidth]{"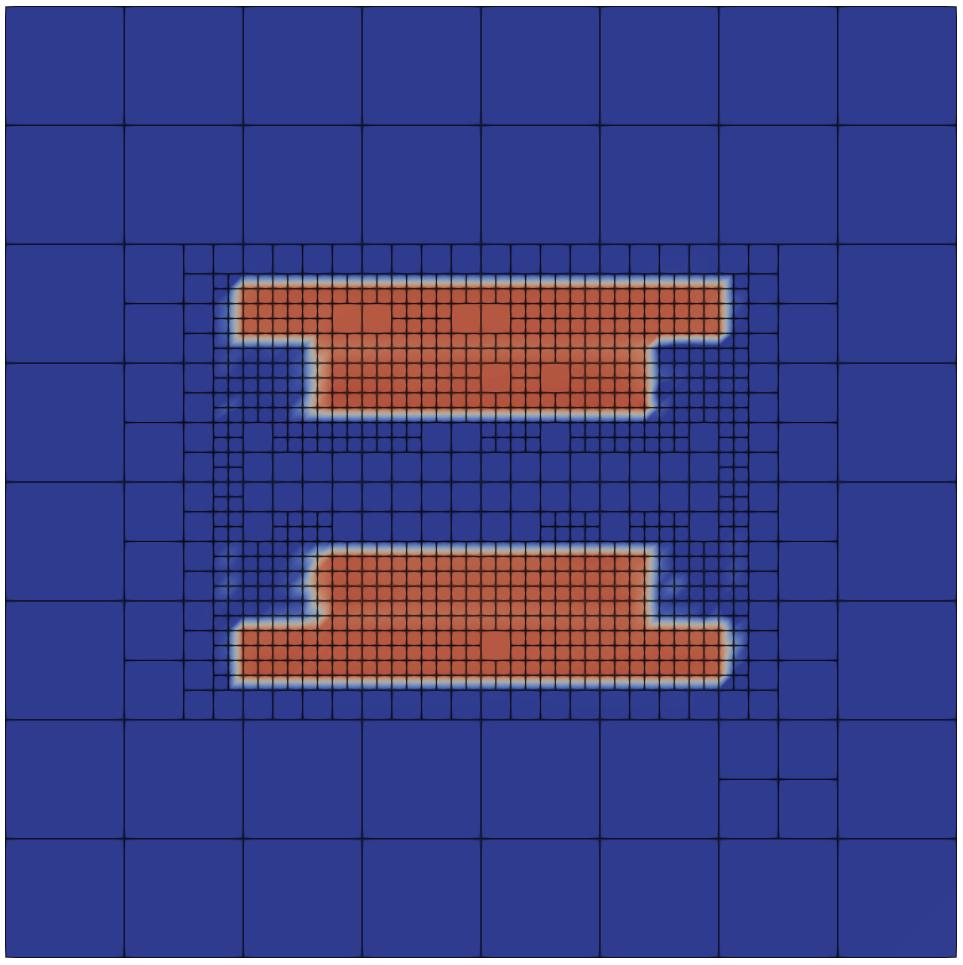"} \\
\includegraphics[width=0.3\linewidth]{"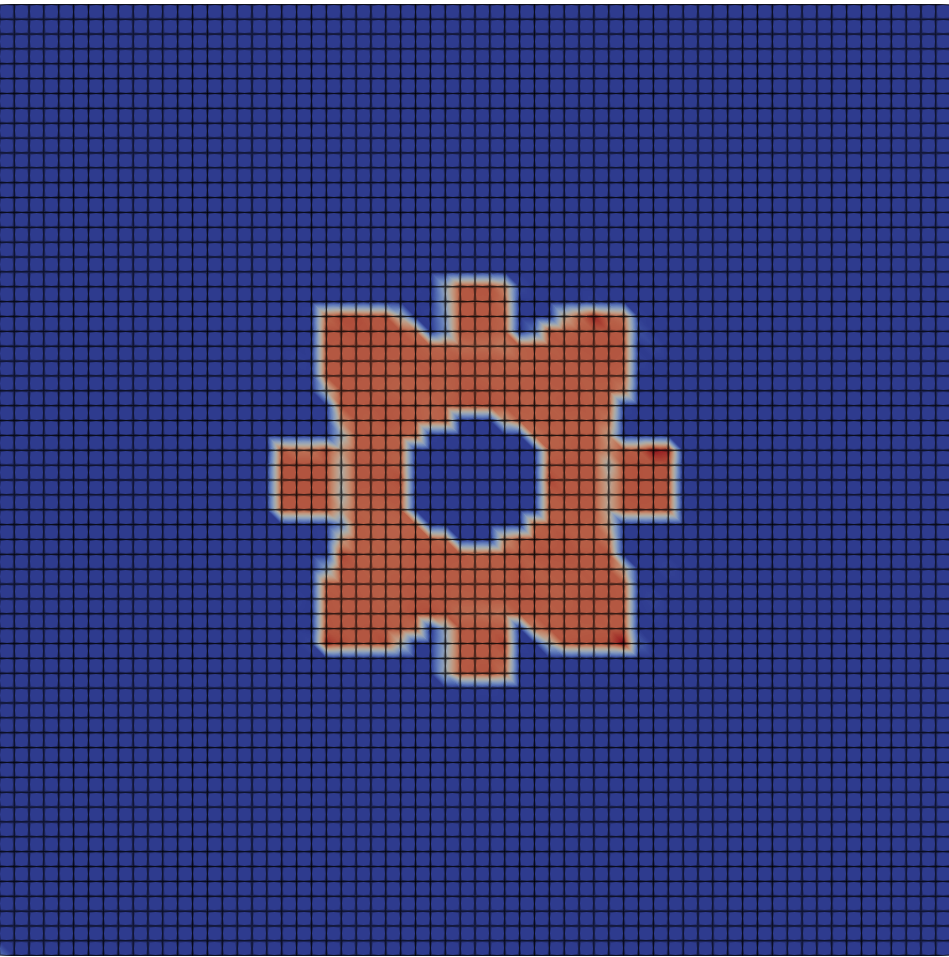"}
& \includegraphics[width=0.3\linewidth]{"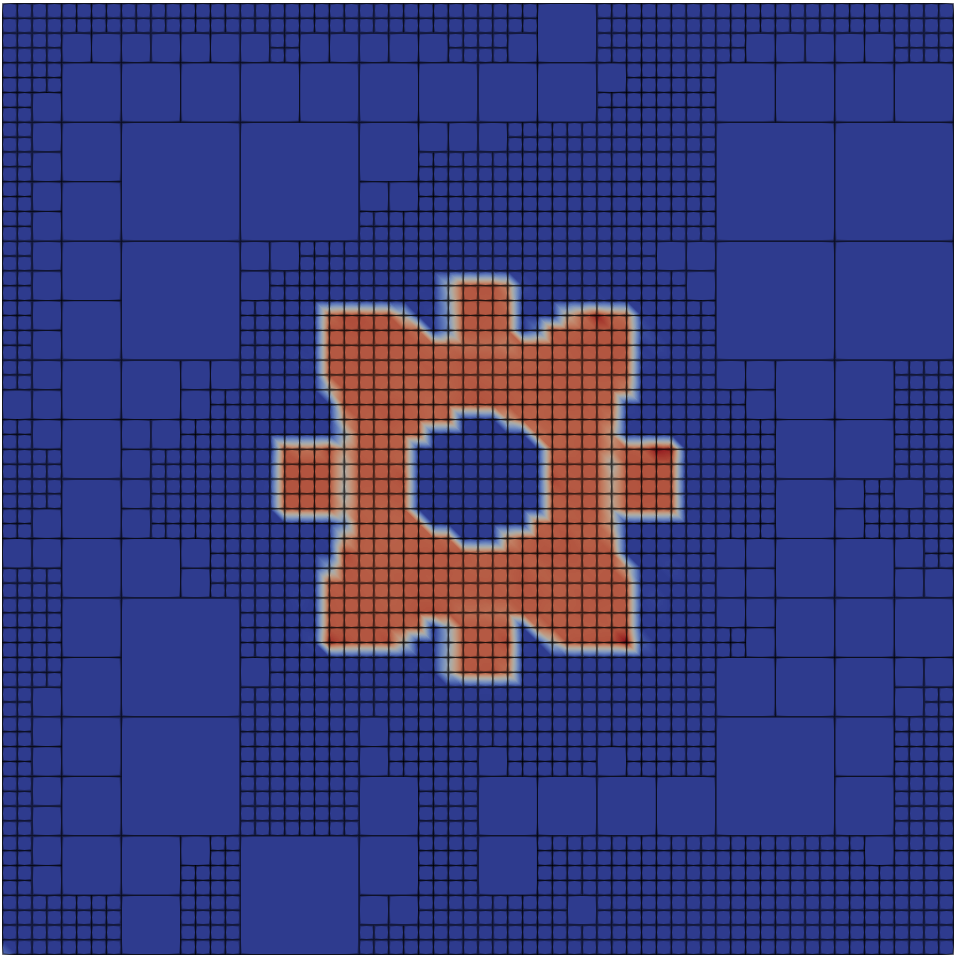"}
&  \includegraphics[width=0.3\linewidth]{"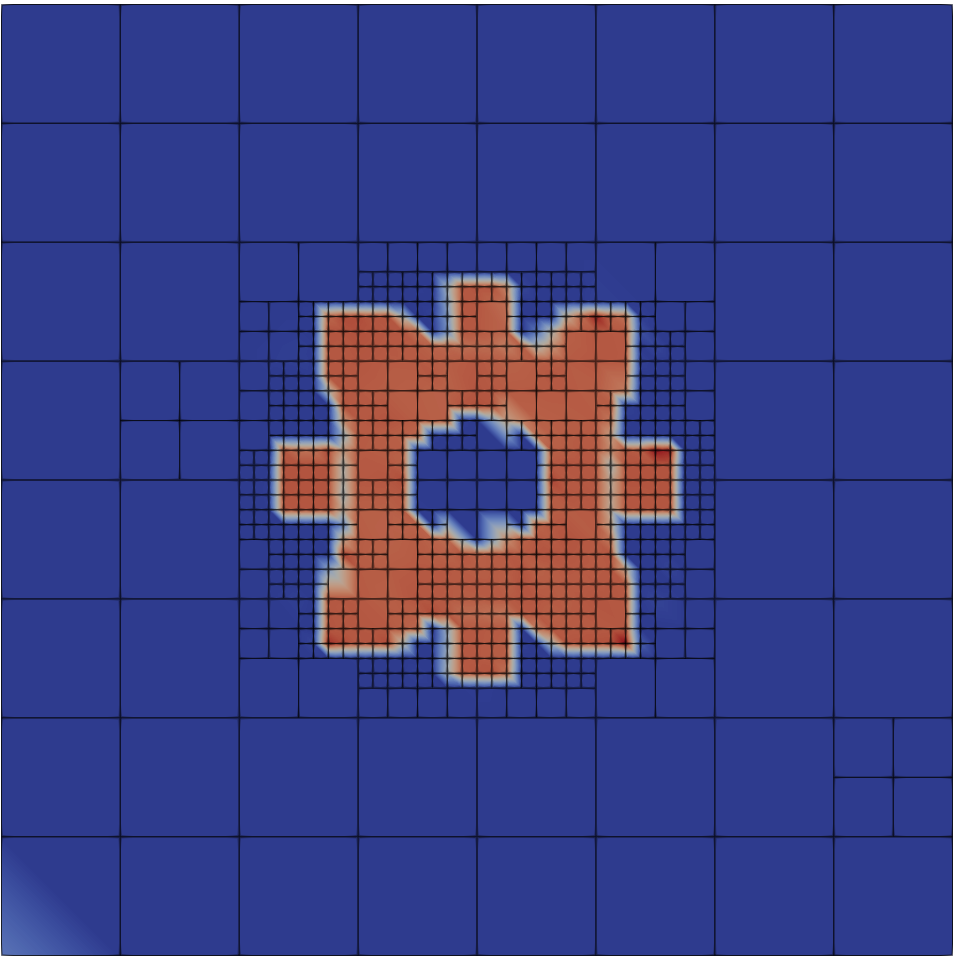"} \\

\texttt{Uniform} & \texttt{BasicAMR} & \texttt{PruningAMR}
\end{tabular}
\caption{Comparison between meshes created by \texttt{Uniform}, \texttt{BasicAMR} ($\tau = 10^{-4}$), and \texttt{PruningAMR} ($T=10^{-4}$, $P=0.075$, $\varepsilon = 10^{-3}$) refinement for the simulated CT INR. Top row: x-slice. Bottom row: y-slice. Each figure shows the result of five iterations of refinement. For each row, the images are visually similar, but \texttt{PruningAMR} uses fewer elements than the other two methods.}
\label{fig:CT5_slices}
\end{figure}

The error and DOFs for each method across six iterations are shown in Figure~\ref{fig:compare_CT5}. We require all methods to perform two uniform refinements first. After these uniform refinements, we see that the \texttt{PruningAMR} curve achieves lower DOFs for a similar level of error to both \texttt{BasicAMR} and \texttt{Uniform}. This difference is reaffirmed in Figure~\ref{fig:CT5_slices}, which shows slices of the simulated CT INR visualization for each of the three refinement methods. The top row shows slices for $x=0$; the bottom row shows slices for $y=0$. Both are taken at the final time, $t=1$. For each row, the visualizations from each method appear similar. 
However, \texttt{PruningAMR} uses fewer elements (and thus, DOFs) than either \texttt{BasicAMR} or \texttt{Uniform}. 
\texttt{PruningAMR} uses 87\% fewer DOFs than the \texttt{Uniform} mesh and 81\% fewer DOFs than the \texttt{BasicAMR} mesh.
\texttt{PruningAMR} also seems to do a better job than \texttt{BasicAMR} at deciding where extra elements are required.  
We expect that the DOFs savings would only further improve with more iterations.
Note that once \texttt{PruningAMR} finishes refining an element, it never refines it again. Thus, \texttt{Uniform} would use increasingly more DOFs than \texttt{PruningAMR} each iteration, with many of these DOFs placed in the background region where there are no features to further resolve.

To demonstrate the utility of our algorithm in 4D, we also show the \texttt{PruningAMR} meshes for three time slices (with $y=0$) in Figure~\ref{fig:CT5_time_slices}. Note that the algorithm chooses a different mesh for each time slice because the object is changing in time, even though the slices are all taken at $y=0$. 

\subsection{Example 3: Experimental dynamic CT INR}
Finally, we consider an INR trained on CT scans from a physical experiment. 
This example contains many more geometric features than the previous examples and includes noise in the region surrounding the object of interest. 
Hence, there are fewer low-detail regions in the INR's domain.

The object of interest in this CT scan is a ``log pile,'' which consists of many layers of strands, or ``logs.'' Each layer has many parallel logs. The layers are rotated 90 degrees relative to each other, so that the logs in one layer are perpendicular to all of the logs in an adjacent layer. The INR used for this example has the same architecture and domain as the INR in Section~\ref{sec:simulation_example}. For more information about the experimental set-up and architecture, see~\citet{CT_INR}. All visualizations for this example are created using ParaView~\citep{Paraview}.

\begin{figure}[h]
  \centering
    % Left: Image
    \begin{minipage}[c]{0.45\textwidth}
        \centering
        \includegraphics[width=\linewidth]{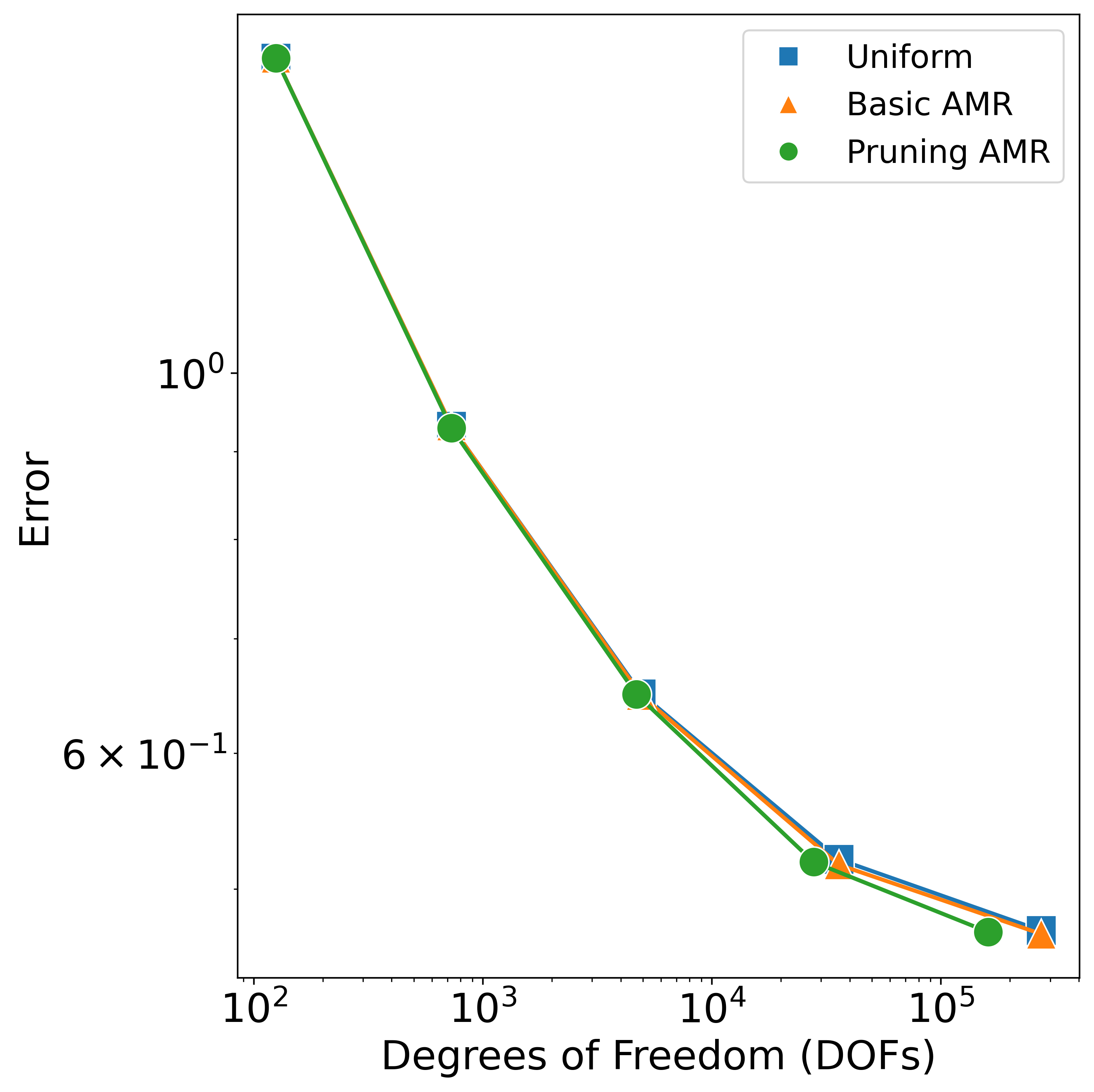}
    \end{minipage}%
    \hfill
    % Right: Table
    \begin{minipage}[c]{0.5\textwidth}
        \centering
    \text{At sixth iteration:}\\
    \text{}\\
        \begin{tabular}{|c|c|c|}
            \hline
         & DOFs & error \\
            \hline
      Uniform   & 274625 & 0.473 \\
      Basic AMR & 274625 & 0.471 \\
      Pruning AMR & 160905 & 0.472\\
            \hline
        \end{tabular}
    \end{minipage}
  \caption{\textbf{Left:} Total error versus number of degrees of freedom are shown for \texttt{Uniform}, \texttt{BasicAMR} ($\tau = 10^{-3}$), and \texttt{PruningAMR} ($T=10^{-3}$, $P=0.1$, $\varepsilon = 10^{-2}$) visualization of a real CT scan of a log pile. All three refinement techniques perform close to uniform refinement until the last iteration, when \texttt{PruningAMR} does marginally better. This example is highly-detailed and would benefit from even more iterations of \texttt{PruningAMR}. \textbf{Right:} Table of DOFs and error values for each algorithm at their final iteration.}
    \label{fig:compare_CT2}
\end{figure}

We applied \texttt{PruningAMR} to the CT INR and compared it to \texttt{Uniform} refinement and \texttt{BasicAMR}, as described in Section~\ref{sec:2D_example}. The results are shown in Figure~\cref{fig:compare_CT2}. For all log pile results, we use the hyperparameters: $T = \tau = 10^{-3}$, $P = 0.1$, $\varepsilon = 10^{-2}$, $K_{max} = 5$, and $n_{ID} = 256$. We use $n_{err} = 256$ for \texttt{PruningAMR} and $n_{err}=512$ for \texttt{BasicAMR}. We use $1048576$ randomly sampled points to compute the root mean squared error for all methods. 

The error and DOFs for each of the three algorithms across iterations 2-5 are shown in Figure~\ref{fig:compare_CT2}. 
\texttt{PruningAMR} uses 41\% fewer DOFs than both \texttt{Uniform} and \texttt{BasicAMR} (which ends up choosing uniform refinement each iteration). 
In this example, \texttt{PruningAMR} only achieves about half the memory savings as in the experimental CT example. 
We believe this reflects the sparsity of low-detail regions in the dataset on which the INR was trained. Thus, both \texttt{PruningAMR} and \texttt{BasicAMR} require many more iterations to get to a small enough scale to take advantage of variable detail across the domain. 

\begin{figure}[H]
\centering
  \begin{tabular}{ccc}
    \includegraphics[width=0.3\linewidth]{"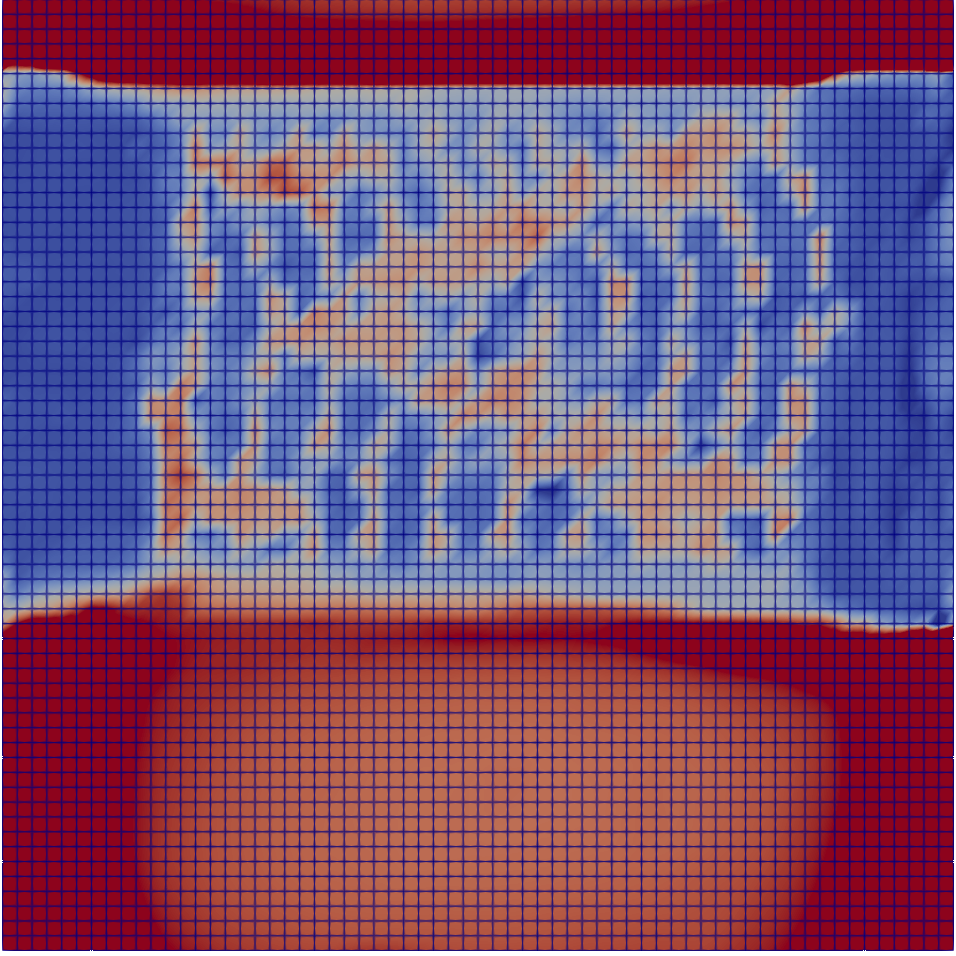"}
    & \includegraphics[width=0.3\linewidth]{"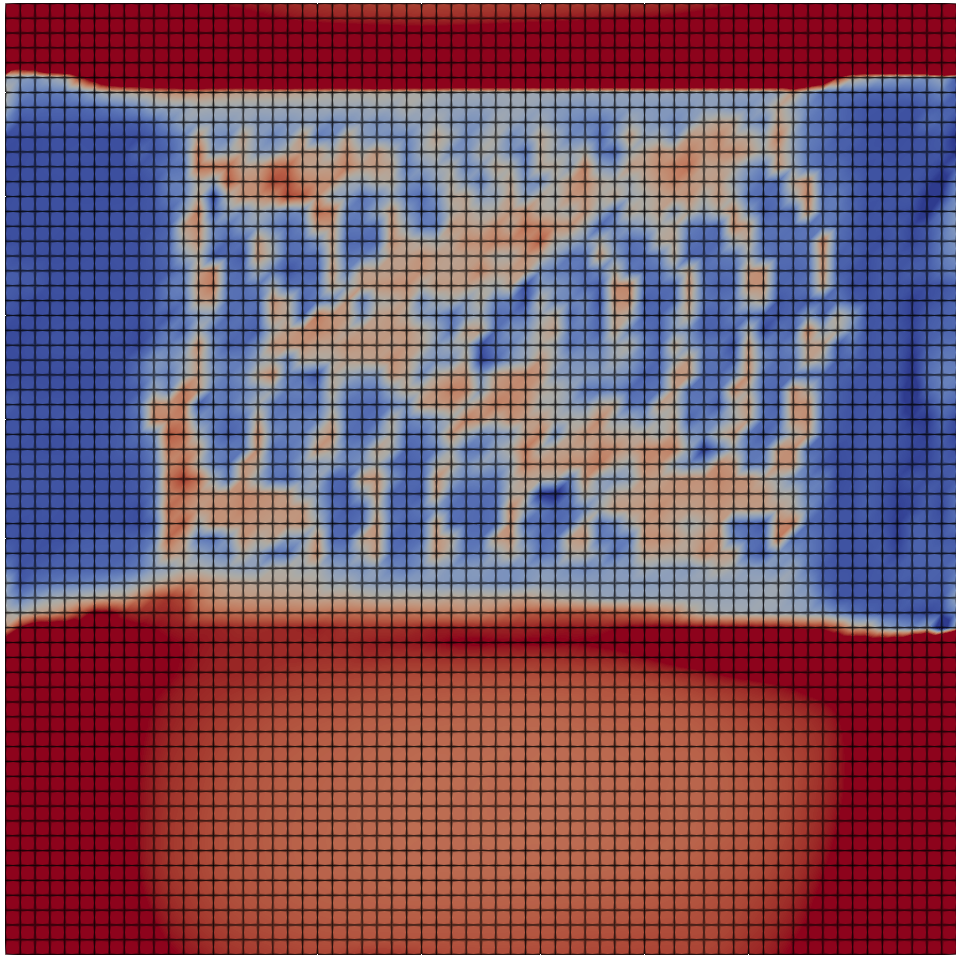"}
    &  \includegraphics[width=0.3\linewidth]{"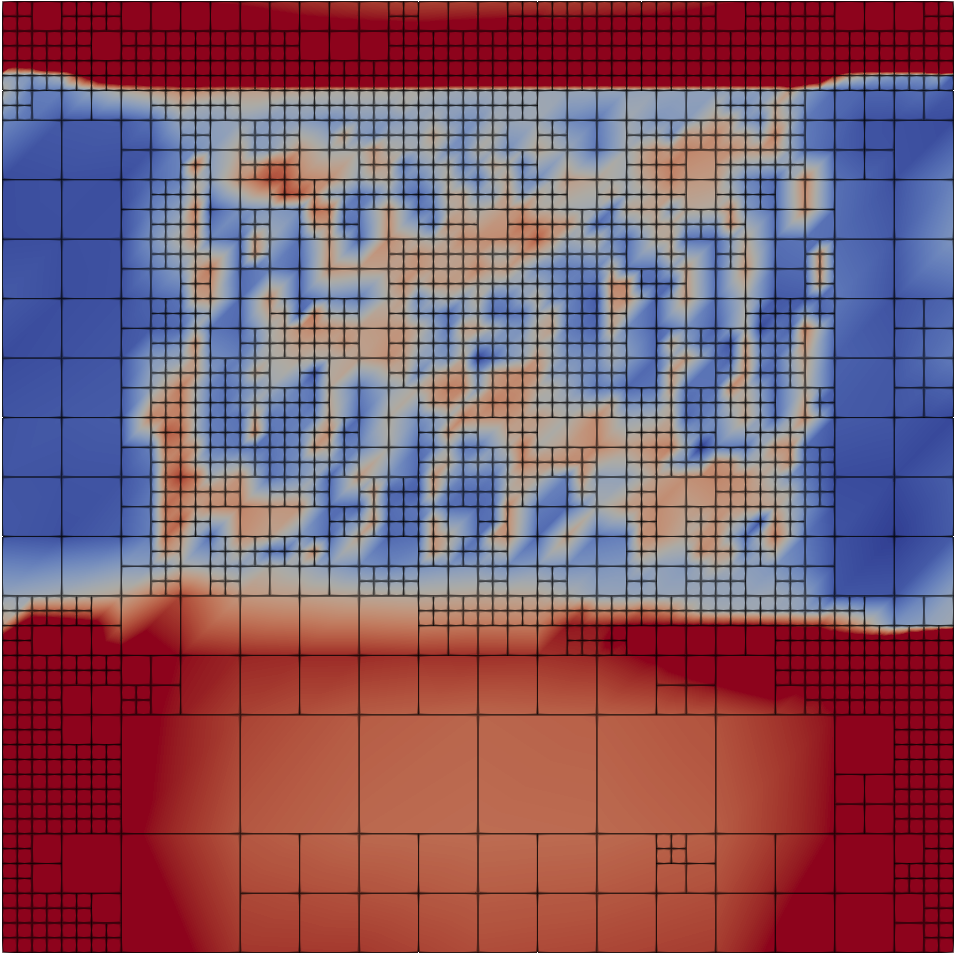"} \\
    \includegraphics[width=0.3\linewidth]{"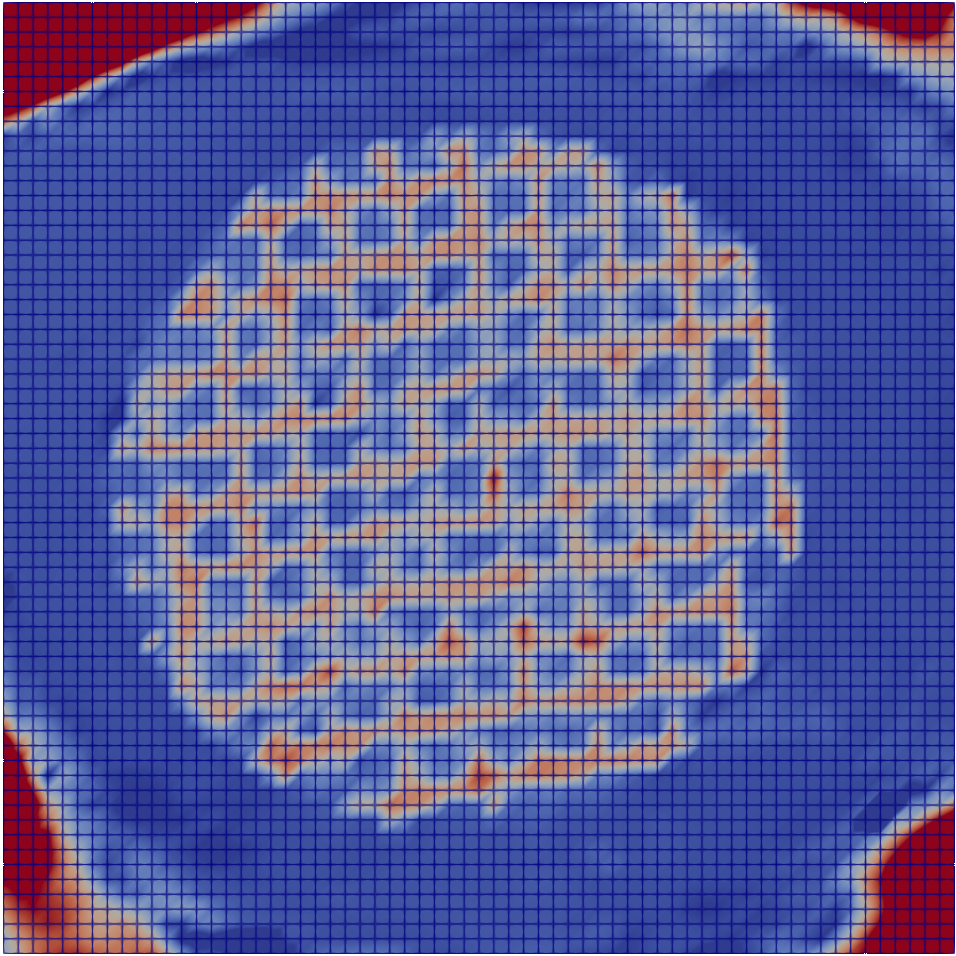"}
    & \includegraphics[width=0.3\linewidth]{"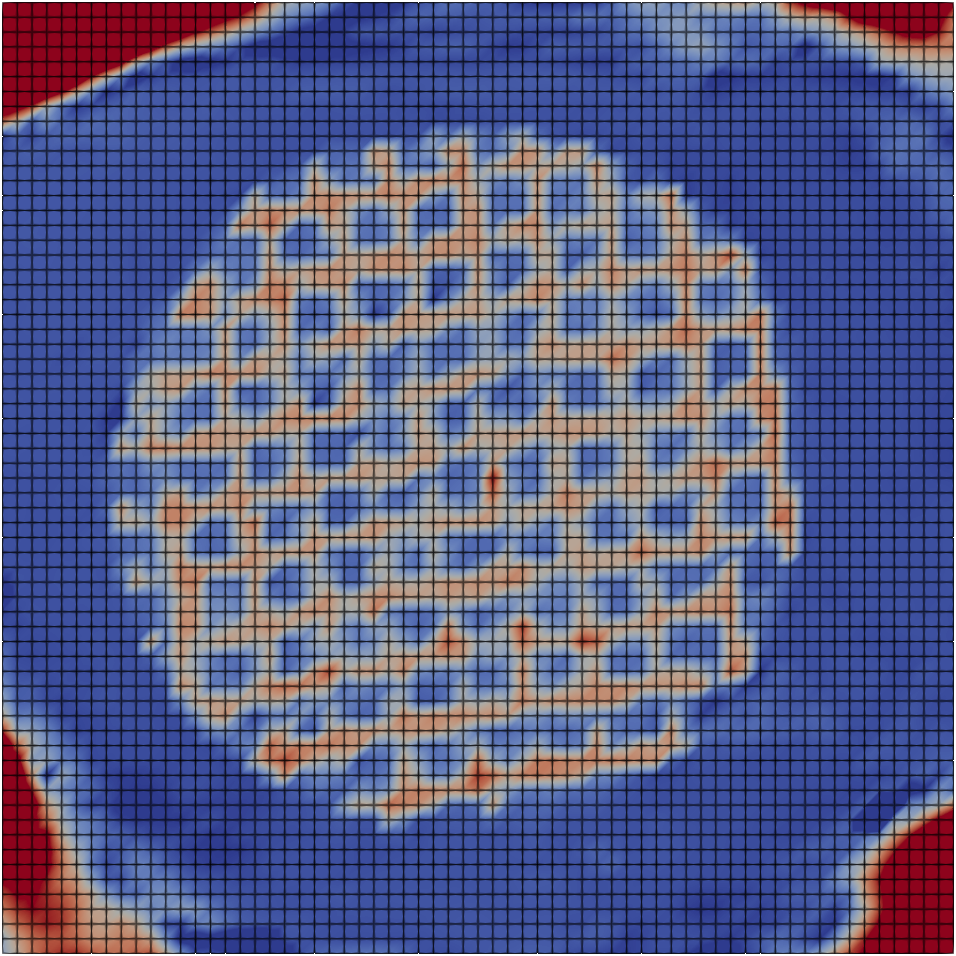"}
    &  \includegraphics[width=0.3\linewidth]{"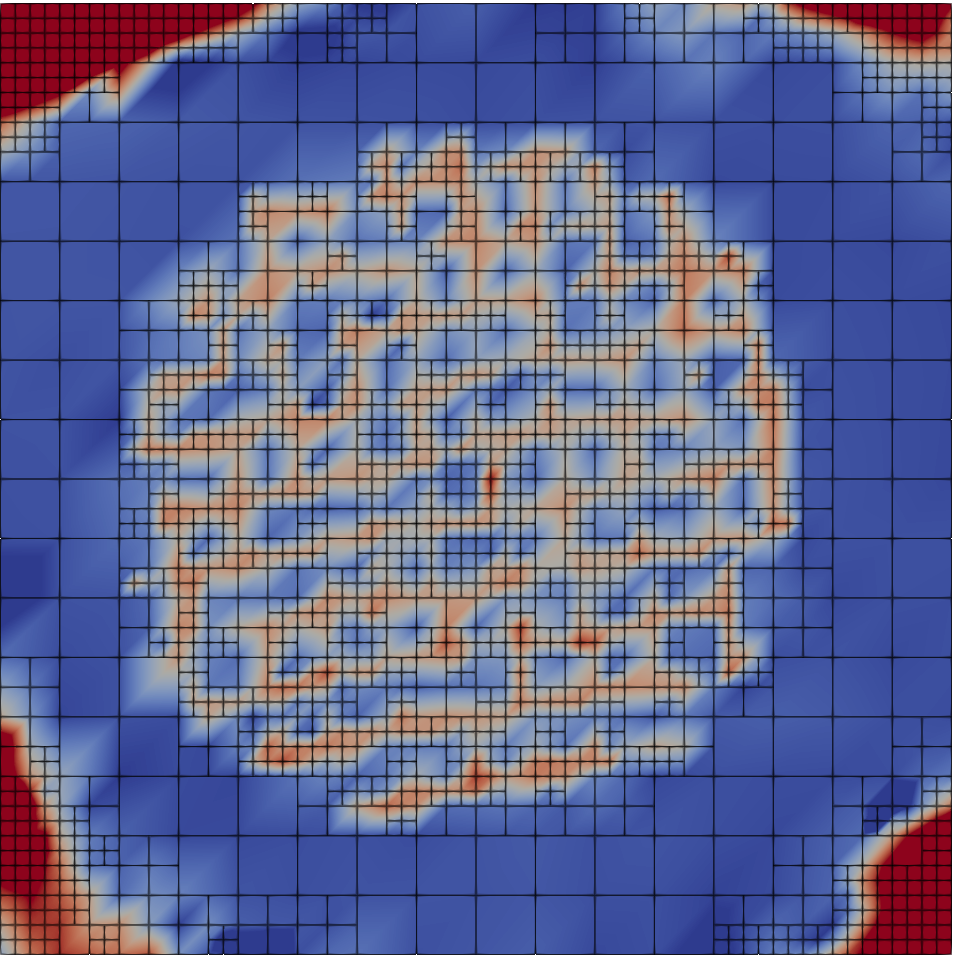"} \\

    \texttt{Uniform} & \texttt{BasicAMR} & \texttt{PruningAMR}
  \end{tabular}
  \caption{Comparison between meshes created by \texttt{Uniform}, \texttt{BasicAMR} ($\tau = 10^{-3}$), and \texttt{PruningAMR} ($T=10^{-3}$, $P=0.1$, $\varepsilon = 10^{-2}$) refinement on log pile CT scan. Top row: x-slice, bottom row: z-slice. Each figure is after five iterations of refinement. For each row, the images look similar but the \texttt{PruningAMR} algorithm uses fewer elements in the less-detailed lower red and circular blue regions, for each row respectively.}
  \label{fig:CT2_slices}
\end{figure}

\pagebreak 
However, differences become more apparent in the fifth iteration. 
For instance, consider Figure~\ref{fig:CT2_slices}, which shows the log pile visualization sliced in the $x$ and $z$ direction at $t=1$ for each of the three algorithms. 
On the top row ($x = 0$) we see that the \texttt{PruningAMR} mesh saves some DOFs in the red region of the figure where there is less variation. 
In the bottom row ($z = 0$), \texttt{PruningAMR} also saves some DOFs in the blue regions around the circular log pile. 
At iteration 5, these savings are smaller than the savings observed in the simulated CT data from Section~\ref{sec:simulation_example}. 
Thus, from this example, we confirm that AMR is more useful for INR visualization if detail is required at a scale for which there are some low-detail regions. 
As with the simulated data, the DOFs savings should only improve with more iterations.

\begin{figure}
\centering
  \begin{tabular}{ccc}
    \includegraphics[width=0.3\linewidth]{"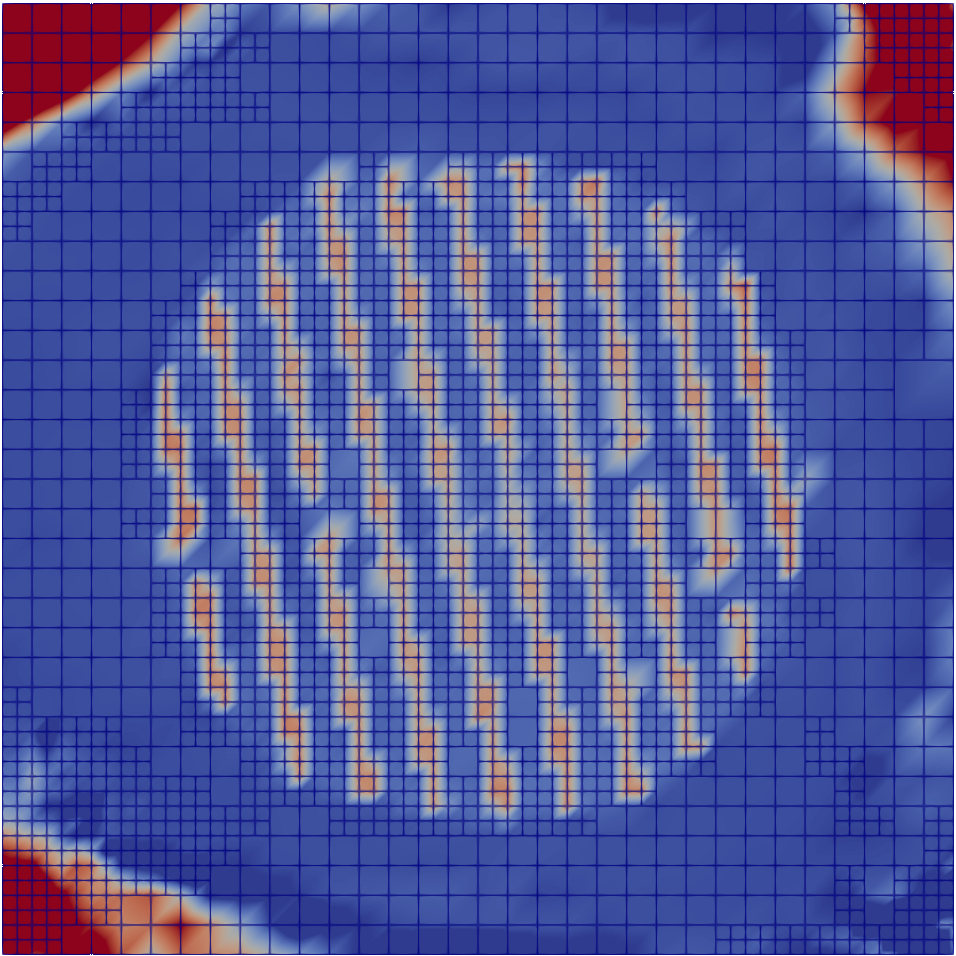"}
    & \includegraphics[width=0.3\linewidth]{"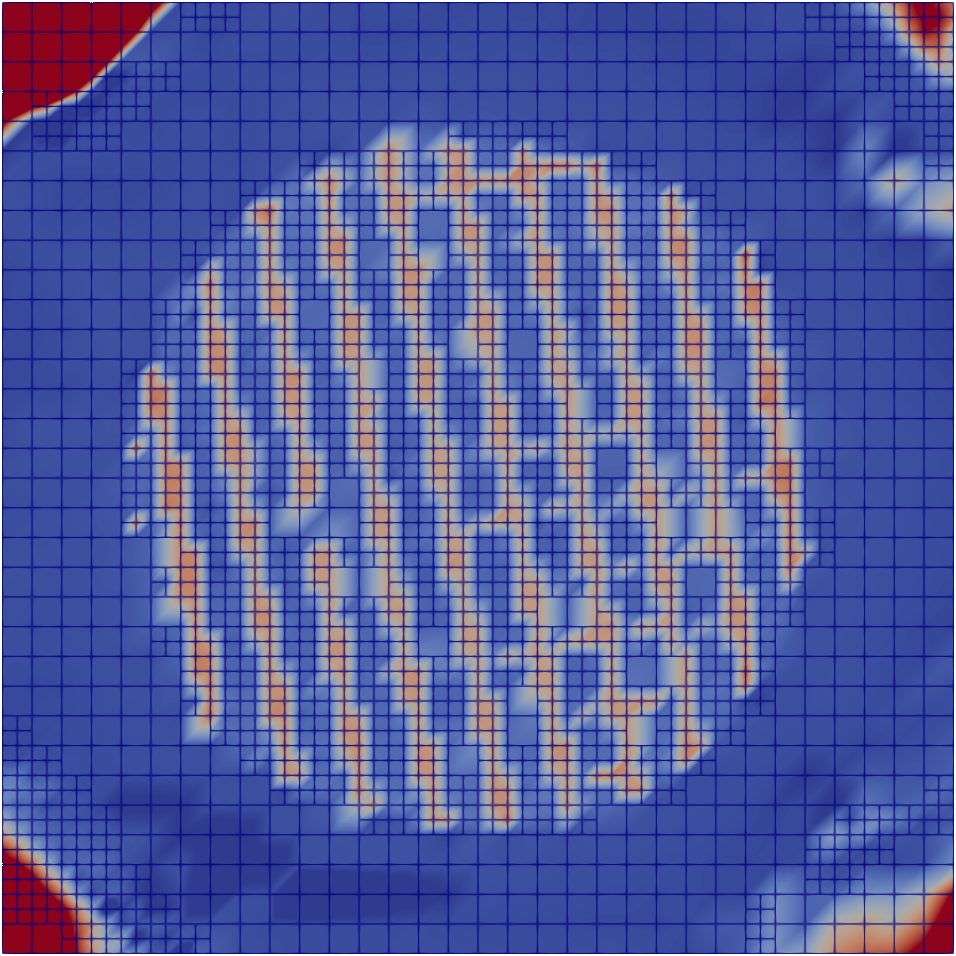"}
    &  \includegraphics[width=0.3\linewidth]{"./CT2_pruned_t=1_zslice.png"} \\
    $t=-1$ & $t=0$ & $t=1$
  \end{tabular}
  \caption{Multiple time slices of the log pile CT INR visualized using \texttt{PruningAMR} ($T=10^{-3}$, $P=0.1$, $\varepsilon = 10^{-2}$), all after five iterations of refinement. From left to right, $t=-1$, $t = 0$, $t = 1$. Notice that the mesh changes with time as the object changes shape.}
  \label{fig:CT2_time_slices}
\end{figure}

Finally, we demonstrate that the mesh changes with time for the experimental INR data. 
Figure~\ref{fig:CT2_time_slices} shows slices of the log pile at $z=0$ for three different times. The mesh adapts to the shape of the object as it deforms in time.

\section{Conclusion}
In this paper we presented an algorithm for finding a variable-resolution visualization of pre-trained implicit neural representations (INRs) with significant memory savings over existing methods. 
The algorithm uses neural network pruning to determine which regions of the INR's domain require higher resolution, then uses adaptive mesh refinement to divide the domain into regions of higher and lower resolution. 
We compared our algorithm to uniform resolution and a simpler variable resolution algorithm; we demonstrated that the \texttt{PruningAMR} algorithm achieves similar error tolerances to these other methods despite using many fewer degrees of freedom. 
We observed that our algorithm is more beneficial for INRs that have their geometric features concentrated in a subset of their input domain.

\subsubsection*{Acknowledgments}
The authors wish to thank thank Frank Wang at LLNL for assistance in the MFEM implementation of the AMR routines.
This work was performed under the auspices of the U.S. Department of Energy by Lawrence Livermore National Laboratory under Contract DE--AC52--07NA27344.  Release number LLNL-JRNL-2013479.

\bibliography{JCP}

\begin{thebibliography}{28}
\expandafter\ifx\csname natexlab\endcsname\relax\def\natexlab#1{#1}\fi
\providecommand{\url}[1]{\texttt{#1}}
\providecommand{\href}[2]{#2}
\providecommand{\path}[1]{#1}
\providecommand{\DOIprefix}{doi:}
\providecommand{\ArXivprefix}{arXiv:}
\providecommand{\URLprefix}{URL: }
\providecommand{\Pubmedprefix}{pmid:}
\providecommand{\doi}[1]{\href{http://dx.doi.org/#1}{\path{#1}}}
\providecommand{\Pubmed}[1]{\href{pmid:#1}{\path{#1}}}
\providecommand{\bibinfo}[2]{#2}
\ifx\xfnm\relax \def\xfnm[#1]{\unskip,\space#1}\fi
%Type = Article
\bibitem[{Ahrens et~al.(2005)Ahrens, Geveci and Law}]{Paraview}
\bibinfo{author}{Ahrens, J.}, \bibinfo{author}{Geveci, B.},
  \bibinfo{author}{Law, C.}, \bibinfo{year}{2005}.
\newblock \bibinfo{title}{Paraview: An end-user tool for large data
  visualization}.
\newblock \bibinfo{journal}{The visualization handbook} \bibinfo{volume}{717}.
%Type = Article
\bibitem[{Anderson et~al.(1999)Anderson, Bai, Bischof, Blackford and
  Demmel}]{LAPACK}
\bibinfo{author}{Anderson, E.}, \bibinfo{author}{Bai, Z.},
  \bibinfo{author}{Bischof, C.}, \bibinfo{author}{Blackford, S.},
  \bibinfo{author}{Demmel, J.}, \bibinfo{year}{1999}.
\newblock \bibinfo{title}{Dongarra j., j. du croz, a. greenbaum, s. hammarling,
  a. mckenney, and d. sorensen}.
\newblock \bibinfo{journal}{LAPACK Users Guide} .
%Type = Article
\bibitem[{Burstedde et~al.(2011)Burstedde, Wilcox and
  Ghattas}]{burstedde2011p4est}
\bibinfo{author}{Burstedde, C.}, \bibinfo{author}{Wilcox, L.C.},
  \bibinfo{author}{Ghattas, O.}, \bibinfo{year}{2011}.
\newblock \bibinfo{title}{p4est: {S}calable algorithms for parallel adaptive
  mesh refinement on forests of octrees}.
\newblock \bibinfo{journal}{SIAM Journal on Scientific Computing}
  \bibinfo{volume}{33}, \bibinfo{pages}{1103--1133}.
%Type = Article
\bibitem[{Chedid and Najjar(1996)}]{chedid1996automatic}
\bibinfo{author}{Chedid, R.}, \bibinfo{author}{Najjar, N.},
  \bibinfo{year}{1996}.
\newblock \bibinfo{title}{Automatic finite-element mesh generation using
  artificial neural networks-{P}art {I}: {P}rediction of mesh density}.
\newblock \bibinfo{journal}{IEEE Transactions on Magnetics}
  \bibinfo{volume}{32}, \bibinfo{pages}{5173--5178}.
%Type = Article
\bibitem[{Chee et~al.(2022)Chee, Flynn, Damle and De~Sa}]{IDPruning}
\bibinfo{author}{Chee, J.}, \bibinfo{author}{Flynn, M.},
  \bibinfo{author}{Damle, A.}, \bibinfo{author}{De~Sa, C.M.},
  \bibinfo{year}{2022}.
\newblock \bibinfo{title}{Model preserving compression for neural networks}.
\newblock \bibinfo{journal}{Advances in Neural Information Processing Systems}
  \bibinfo{volume}{35}, \bibinfo{pages}{38060--38074}.
%Type = Article
\bibitem[{Dyck et~al.(1992)Dyck, Lowther and McFee}]{dyck1992determining}
\bibinfo{author}{Dyck, D.}, \bibinfo{author}{Lowther, D.},
  \bibinfo{author}{McFee, S.}, \bibinfo{year}{1992}.
\newblock \bibinfo{title}{Determining an approximate finite element mesh
  density using neural network techniques}.
\newblock \bibinfo{journal}{IEEE transactions on magnetics}
  \bibinfo{volume}{28}, \bibinfo{pages}{1767--1770}.
%Type = Article
\bibitem[{Dzanic et~al.(2024)Dzanic, Mittal, Kim, Yang, Petrides, Keith and
  Anderson}]{dzanic2024dynamo}
\bibinfo{author}{Dzanic, T.}, \bibinfo{author}{Mittal, K.},
  \bibinfo{author}{Kim, D.}, \bibinfo{author}{Yang, J.},
  \bibinfo{author}{Petrides, S.}, \bibinfo{author}{Keith, B.},
  \bibinfo{author}{Anderson, R.}, \bibinfo{year}{2024}.
\newblock \bibinfo{title}{Dynamo: Multi-agent reinforcement learning for
  dynamic anticipatory mesh optimization with applications to hyperbolic
  conservation laws}.
\newblock \bibinfo{journal}{Journal of Computational Physics}
  \bibinfo{volume}{506}, \bibinfo{pages}{112924}.
%Type = Article
\bibitem[{Gillette et~al.(2024)Gillette, Keith and
  Petrides}]{gillette2024learning}
\bibinfo{author}{Gillette, A.}, \bibinfo{author}{Keith, B.},
  \bibinfo{author}{Petrides, S.}, \bibinfo{year}{2024}.
\newblock \bibinfo{title}{Learning robust marking policies for adaptive mesh
  refinement}.
\newblock \bibinfo{journal}{SIAM Journal on Scientific Computing}
  \bibinfo{volume}{46}, \bibinfo{pages}{A264--A289}.
%Type = Misc
\bibitem[{GLVis()}]{glvis-web}
GLVis, .
\newblock \bibinfo{title}{{GLVis}: {OpenGL} {F}inite {E}lement {V}isualization
  {T}ool}.
\newblock \bibinfo{howpublished}{\url{glvis.org}}.
\newblock \DOIprefix\doi{10.11578/dc.20171025.1249}.
%Type = Article
\bibitem[{Karniadakis et~al.(2021)Karniadakis, Kevrekidis, Lu, Perdikaris, Wang
  and Yang}]{karniadakis2021physics}
\bibinfo{author}{Karniadakis, G.E.}, \bibinfo{author}{Kevrekidis, I.G.},
  \bibinfo{author}{Lu, L.}, \bibinfo{author}{Perdikaris, P.},
  \bibinfo{author}{Wang, S.}, \bibinfo{author}{Yang, L.}, \bibinfo{year}{2021}.
\newblock \bibinfo{title}{Physics-informed machine learning}.
\newblock \bibinfo{journal}{Nature Reviews Physics} \bibinfo{volume}{3},
  \bibinfo{pages}{422--440}.
%Type = Article
\bibitem[{Lee et~al.(2018)Lee, Ajanthan and Torr}]{lee2018snip}
\bibinfo{author}{Lee, N.}, \bibinfo{author}{Ajanthan, T.},
  \bibinfo{author}{Torr, P.H.}, \bibinfo{year}{2018}.
\newblock \bibinfo{title}{Snip: {S}ingle-shot network pruning based on
  connection sensitivity}.
\newblock \bibinfo{journal}{arXiv preprint arXiv:1810.02340} .
%Type = Article
\bibitem[{Li et~al.(2016)Li, Kadav, Durdanovic, Samet and Graf}]{li2016pruning}
\bibinfo{author}{Li, H.}, \bibinfo{author}{Kadav, A.},
  \bibinfo{author}{Durdanovic, I.}, \bibinfo{author}{Samet, H.},
  \bibinfo{author}{Graf, H.P.}, \bibinfo{year}{2016}.
\newblock \bibinfo{title}{Pruning filters for efficient convnets}.
\newblock \bibinfo{journal}{arXiv preprint arXiv:1608.08710} .
%Type = Article
\bibitem[{Liebenwein et~al.(2019)Liebenwein, Baykal, Lang, Feldman and
  Rus}]{liebenwein2019provable}
\bibinfo{author}{Liebenwein, L.}, \bibinfo{author}{Baykal, C.},
  \bibinfo{author}{Lang, H.}, \bibinfo{author}{Feldman, D.},
  \bibinfo{author}{Rus, D.}, \bibinfo{year}{2019}.
\newblock \bibinfo{title}{Provable filter pruning for efficient neural
  networks}.
\newblock \bibinfo{journal}{arXiv preprint arXiv:1911.07412} .
%Type = Article
\bibitem[{Liu et~al.(2018)Liu, Sun, Zhou, Huang and
  Darrell}]{liu2018rethinking}
\bibinfo{author}{Liu, Z.}, \bibinfo{author}{Sun, M.}, \bibinfo{author}{Zhou,
  T.}, \bibinfo{author}{Huang, G.}, \bibinfo{author}{Darrell, T.},
  \bibinfo{year}{2018}.
\newblock \bibinfo{title}{Rethinking the value of network pruning}.
\newblock \bibinfo{journal}{arXiv preprint arXiv:1810.05270} .
%Type = Article
\bibitem[{Martel et~al.(2021)Martel, Lindell, Lin, Chan, Monteiro and
  Wetzstein}]{martel2021acorn}
\bibinfo{author}{Martel, J.N.}, \bibinfo{author}{Lindell, D.B.},
  \bibinfo{author}{Lin, C.Z.}, \bibinfo{author}{Chan, E.R.},
  \bibinfo{author}{Monteiro, M.}, \bibinfo{author}{Wetzstein, G.},
  \bibinfo{year}{2021}.
\newblock \bibinfo{title}{{ACORN}: {A}daptive coordinate networks for neural
  representation}.
\newblock \bibinfo{journal}{ACM Trans. Graph. (SIGGRAPH)} .
%Type = Misc
\bibitem[{MFEM()}]{mfem-web}
MFEM, .
\newblock \bibinfo{title}{{MFEM}: {M}odular {F}inite {E}lement {M}ethods
  {[Software]}}.
\newblock \bibinfo{howpublished}{\url{mfem.org}}.
\newblock \DOIprefix\doi{10.11578/dc.20171025.1248}.
%Type = Article
\bibitem[{Mildenhall et~al.(2021)Mildenhall, Srinivasan, Tancik, Barron,
  Ramamoorthi and Ng}]{mildenhall2021nerf}
\bibinfo{author}{Mildenhall, B.}, \bibinfo{author}{Srinivasan, P.P.},
  \bibinfo{author}{Tancik, M.}, \bibinfo{author}{Barron, J.T.},
  \bibinfo{author}{Ramamoorthi, R.}, \bibinfo{author}{Ng, R.},
  \bibinfo{year}{2021}.
\newblock \bibinfo{title}{Nerf: {R}epresenting scenes as neural radiance fields
  for view synthesis}.
\newblock \bibinfo{journal}{Communications of the ACM} \bibinfo{volume}{65},
  \bibinfo{pages}{99--106}.
%Type = Article
\bibitem[{Mitchell(2013)}]{mitchell2013collection}
\bibinfo{author}{Mitchell, W.F.}, \bibinfo{year}{2013}.
\newblock \bibinfo{title}{A collection of {2D} elliptic problems for testing
  adaptive grid refinement algorithms}.
\newblock \bibinfo{journal}{Applied mathematics and computation}
  \bibinfo{volume}{220}, \bibinfo{pages}{350--364}.
%Type = Article
\bibitem[{Mohan et~al.(2024)Mohan, Ferrucci, Divin, Stevenson and Kim}]{CT_INR}
\bibinfo{author}{Mohan, K.A.}, \bibinfo{author}{Ferrucci, M.},
  \bibinfo{author}{Divin, C.}, \bibinfo{author}{Stevenson, G.A.},
  \bibinfo{author}{Kim, H.}, \bibinfo{year}{2024}.
\newblock \bibinfo{title}{Distributed stochastic optimization of a neural
  representation network for time-space tomography reconstruction}.
\newblock \bibinfo{journal}{arXiv preprint arXiv:2404.19075} .
%Type = Article
\bibitem[{Mussay et~al.(2019)Mussay, Osadchy, Braverman, Zhou and
  Feldman}]{mussay2019data}
\bibinfo{author}{Mussay, B.}, \bibinfo{author}{Osadchy, M.},
  \bibinfo{author}{Braverman, V.}, \bibinfo{author}{Zhou, S.},
  \bibinfo{author}{Feldman, D.}, \bibinfo{year}{2019}.
\newblock \bibinfo{title}{Data-independent neural pruning via coresets}.
\newblock \bibinfo{journal}{arXiv preprint arXiv:1907.04018} .
%Type = Inproceedings
\bibitem[{Peterka et~al.(2023)Peterka, Lenz, Grindeanu and
  Mahadevan}]{peterka2023towards}
\bibinfo{author}{Peterka, T.}, \bibinfo{author}{Lenz, D.},
  \bibinfo{author}{Grindeanu, I.}, \bibinfo{author}{Mahadevan, V.S.},
  \bibinfo{year}{2023}.
\newblock \bibinfo{title}{Towards adaptive refinement for multivariate
  functional approximation of scientific data}, in: \bibinfo{booktitle}{2023
  IEEE 13th Symposium on Large Data Analysis and Visualization (LDAV)},
  \bibinfo{organization}{IEEE}. pp. \bibinfo{pages}{32--41}.
%Type = Article
\bibitem[{Raissi et~al.(2019)Raissi, Perdikaris and Karniadakis}]{PINN}
\bibinfo{author}{Raissi, M.}, \bibinfo{author}{Perdikaris, P.},
  \bibinfo{author}{Karniadakis, G.E.}, \bibinfo{year}{2019}.
\newblock \bibinfo{title}{Physics-informed neural networks: A deep learning
  framework for solving forward and inverse problems involving nonlinear
  partial differential equations}.
\newblock \bibinfo{journal}{Journal of Computational physics}
  \bibinfo{volume}{378}, \bibinfo{pages}{686--707}.
%Type = Article
\bibitem[{Sitzmann et~al.(2020)Sitzmann, Martel, Bergman, Lindell and
  Wetzstein}]{sitzmann2020implicit}
\bibinfo{author}{Sitzmann, V.}, \bibinfo{author}{Martel, J.},
  \bibinfo{author}{Bergman, A.}, \bibinfo{author}{Lindell, D.},
  \bibinfo{author}{Wetzstein, G.}, \bibinfo{year}{2020}.
\newblock \bibinfo{title}{Implicit neural representations with periodic
  activation functions}.
\newblock \bibinfo{journal}{Advances in neural information processing systems}
  \bibinfo{volume}{33}, \bibinfo{pages}{7462--7473}.
%Type = Article
\bibitem[{Tancik et~al.(2020)Tancik, Srinivasan, Mildenhall, Fridovich-Keil,
  Raghavan, Singhal, Ramamoorthi, Barron and Ng}]{tancik2020fourier}
\bibinfo{author}{Tancik, M.}, \bibinfo{author}{Srinivasan, P.},
  \bibinfo{author}{Mildenhall, B.}, \bibinfo{author}{Fridovich-Keil, S.},
  \bibinfo{author}{Raghavan, N.}, \bibinfo{author}{Singhal, U.},
  \bibinfo{author}{Ramamoorthi, R.}, \bibinfo{author}{Barron, J.},
  \bibinfo{author}{Ng, R.}, \bibinfo{year}{2020}.
\newblock \bibinfo{title}{Fourier features let networks learn high frequency
  functions in low dimensional domains}.
\newblock \bibinfo{journal}{Advances in neural information processing systems}
  \bibinfo{volume}{33}, \bibinfo{pages}{7537--7547}.
%Type = Inproceedings
\bibitem[{Wang et~al.(2020)Wang, Marshak, Usher, Burstedde, Knoll, Heister and
  Johnson}]{wang2020cpu}
\bibinfo{author}{Wang, F.}, \bibinfo{author}{Marshak, N.},
  \bibinfo{author}{Usher, W.}, \bibinfo{author}{Burstedde, C.},
  \bibinfo{author}{Knoll, A.}, \bibinfo{author}{Heister, T.},
  \bibinfo{author}{Johnson, C.R.}, \bibinfo{year}{2020}.
\newblock \bibinfo{title}{{CPU} ray tracing of tree-based adaptive mesh
  refinement data}, in: \bibinfo{booktitle}{Computer graphics forum},
  \bibinfo{organization}{Wiley Online Library}. pp. \bibinfo{pages}{1--12}.
%Type = Article
\bibitem[{Wurster et~al.(2023)Wurster, Xiong, Shen, Guo and
  Peterka}]{wurster2023adaptively}
\bibinfo{author}{Wurster, S.W.}, \bibinfo{author}{Xiong, T.},
  \bibinfo{author}{Shen, H.W.}, \bibinfo{author}{Guo, H.},
  \bibinfo{author}{Peterka, T.}, \bibinfo{year}{2023}.
\newblock \bibinfo{title}{Adaptively placed multi-grid scene representation
  networks for large-scale data visualization}.
\newblock \bibinfo{journal}{IEEE Transactions on Visualization and Computer
  Graphics} .
%Type = Article
\bibitem[{Xu et~al.(2023)Xu, Sun, Huang, Guo, Yang and Ju}]{NSPINN}
\bibinfo{author}{Xu, S.}, \bibinfo{author}{Sun, Z.}, \bibinfo{author}{Huang,
  R.}, \bibinfo{author}{Guo, D.}, \bibinfo{author}{Yang, G.},
  \bibinfo{author}{Ju, S.}, \bibinfo{year}{2023}.
\newblock \bibinfo{title}{A practical approach to flow field reconstruction
  with sparse or incomplete data through physics informed neural network}.
\newblock \bibinfo{journal}{Acta Mechanica Sinica} \bibinfo{volume}{39},
  \bibinfo{pages}{322302}.
%Type = Misc
\bibitem[{Zvonek and Gillette(2025)}]{PruningAMR_code}
\bibinfo{author}{Zvonek, J.E.}, \bibinfo{author}{Gillette, A.K.},
  \bibinfo{year}{2025}.
\newblock \bibinfo{title}{Pruningamr}.
\newblock \bibinfo{howpublished}{[Computer Software]
  \url{https://doi.org/10.11578/dc.20250924.3}}.
\newblock \URLprefix \url{https://doi.org/10.11578/dc.20250924.3},
  \DOIprefix\doi{10.11578/dc.20250924.3}.

\end{thebibliography}
\bibliographystyle{elsarticle-harv}

\end{document}